%% file: main.tex
\documentclass[10pt,twocolumn,letterpaper]{article}

\usepackage[pagenumbers]{style}
\definecolor{blue}{rgb}{0.21,0.49,0.74}
\usepackage[pagebackref,breaklinks,colorlinks,allcolors=blue]{hyperref}
\definecolor{pinkcolor}{RGB}{255, 105, 180}  

\usepackage{xcolor}
\usepackage{appendix}
\usepackage{breqn}
\usepackage{caption}
\usepackage{csquotes}
\usepackage{xspace}
\usepackage{mathtools}
\usepackage{amsmath}
\usepackage{graphicx}
\usepackage{booktabs}
\usepackage{multirow}
\usepackage{amsfonts} 
\usepackage{graphicx}
\usepackage{adjustbox}
\usepackage{tabularx}
\usepackage{tablefootnote}
\usepackage{float}
\usepackage{colortbl}
\usepackage{arydshln}

\newcommand*{\competeAgainst}{21\xspace}
\newcommand*{\numberOfTasks}{4\xspace}
\newcommand*{\foundationModels}{3\xspace}

\newcommand*{\TotalDatasets}{46\xspace}

\newcommand*{\ViTgiantMemReduction}{52.63\%\xspace}
\newcommand*{\ViTgiantTimeReduction}{43.0\%\xspace}
\newcommand*{\NumAdaptationMethods}{17\xspace}

\newcommand*{\apla}{\textsc{APLA}\xspace}
\newcommand*{\wo}{$W_O$\xspace}
\newcommand*{\wq}{$W_Q$\xspace}
\newcommand*{\wk}{$W_K$\xspace}
\newcommand*{\wv}{$W_V$\xspace}
\newcommand{\wfcone}{$W_{FC_1}$}
\newcommand{\wfctwo}{$W_{FC_2}$}
\newcommand*{\finetune}{\textsc{Finetune}\xspace}
\newcommand*{\linear}{\textsc{Linear}\xspace}
\newcommand*{\mlp}{\textsc{MLP}\xspace}
\newcommand*{\mlpK}{\textsc{MLP-}$k$\xspace}
\newcommand*{\partialadapt}{\textsc{Partial}\xspace}
\newcommand*{\partialadaptK}{\textsc{Partial-}$k$\xspace}
\newcommand*{\bias}{\textsc{BitFit}\xspace}
\newcommand*{\adapter}{\textsc{Adapter}\xspace}
\newcommand*{\adaptformer}{\textsc{AdaptFormer}\xspace}
\newcommand*{\vpt}{\textsc{VPT}\xspace}
\newcommand*{\vptshallow}{\textsc{VPT-shallow}\xspace}
\newcommand*{\vptdeep}{\textsc{VPT-Deep}\xspace}
\newcommand*{\ssf}{\textsc{SSF}\xspace}
\newcommand*{\lora}{\textsc{LoRA}\xspace}
\newcommand*{\spt}{\textsc{SPT}\xspace}
\newcommand*{\sptadapter}{\textsc{SPT-Adapter}\xspace}
\newcommand*{\sptlora}{\textsc{SPT-LoRA}\xspace}
\newcommand*{\fact}{\textsc{FacT}\xspace}
\newcommand*{\facttk}{\textsc{FacT-TK}\xspace}
\newcommand*{\facttt}{\textsc{FacT-TT}\xspace}
\newcommand*{\arc}{\textsc{ARC}\xspace}
\newcommand*{\rlrr}{\textsc{RLRR}\xspace}
\newcommand*{\gps}{\textsc{GPS}\xspace}
\newcommand*{\noah}{\textsc{NOAH}\xspace}
\newcommand*{\consolidator}{\textsc{Consolidator}\xspace}
\newcommand*{\etwovpt}{\textsc{E\textsuperscript{2}VPT}\xspace}

\newcommand*{\vits}{\textsc{V}i\textsc{T-S}\xspace}
\newcommand*{\vitb}{\textsc{V}i\textsc{T-B}\xspace}
\newcommand*{\vitl}{\textsc{V}i\textsc{T-L}\xspace}
\newcommand*{\vitg}{\textsc{V}i\textsc{T-}g\xspace}
\newcommand*{\dinovtwo}{\textsc{DinoV2}\xspace}
\newcommand*{\openclip}{\textsc{OpenCLIP}\xspace}
\newcommand*{\imagenettwentyonek}{\textsc{ImageNet}-21K\xspace}
\newcommand*{\imagenetOneK}{\textsc{ImageNet}-1K\xspace}

\newcommand*{\vtab}{VTAB\xspace}
\newcommand*{\vtabOneK}{VTAB-1k\xspace}
\newcommand*{\cls}{\textsc{[CLS]}\xspace}
\newcommand*{\coco}{MS COCO\xspace}
\newcommand*{\ade}{ADE20K\xspace}

\newcommand{\rotateN}{90}

\newcommand{\avgcolspace}{5pt}

\newcommand*{\birds}{{Birds}\xspace}
\newcommand*{\cars}{{Cars}\xspace}
\newcommand*{\aid}{{AID}\xspace}
\newcommand*{\isic}{{ISIC}\xspace}
\newcommand*{\average}{{Average}\xspace}

\newcommand*{\vtabnatural}{{Natrl.}\xspace}
\newcommand*{\vtabspecialized}{{Spec.}\xspace}
\newcommand*{\vtabstructured}{{Struc.}\xspace}

\newcommand{\tSNEPanelWidth}{0.24\columnwidth}
\newcommand{\tSNEFigureWidth}{0.48\textwidth}
\newcommand{\tSNEColSpace}{0.8 pt}
\newcommand{\tSNETitle}[1]{\tiny{#1}}

\makeatletter
\def\adl@drawiv#1#2#3{%
        \xleaders#3{#2.3\@tempdimb #1{1}#2.3\@tempdimb}%
                #2\z@ plus1fil minus1fil\relax
        \hskip.5\tabcolsep}
\newcommand{\cdashlinelr}[1]{%
  \noalign{\vskip\aboverulesep
           \global\let\@dashdrawstore\adl@draw
           \global\let\adl@draw\adl@drawiv}
  \cdashline{#1}
  \noalign{\global\let\adl@draw\@dashdrawstore
           \vskip\belowrulesep}}
\makeatother

\title{
APLA: A Simple Adaptation Method for Vision Transformers
}

\author{
Moein Sorkhei\textsuperscript{1,2}
\thanks{Corresponding author: Moein Sorkhei \texttt{<sorkhei@kth.se>}} \quad 
Emir Konuk\textsuperscript{1,2} \quad 
Kevin Smith\textsuperscript{1,2}
\thanks{Equal contribution.}
\quad 
Christos Matsoukas\textsuperscript{1,2} \footnotemark[2]
\\
\textsuperscript{1} KTH Royal Institute of Technology, Stockholm, Sweden\\ \textsuperscript{2}Science for Life Laboratory, Stockholm, Sweden
}

\begin{document}
\maketitle

\input{0_abstract}

\input{1_introduction}
\input{2_related}
\input{3_methods}
\input{4_experimental_setup}
\input{5_experiments}
\input{6_discussion}
\input{7_conclusion}
\input{8_acknowledgements}

{
    \small
    \bibliographystyle{ieeenat_fullname}
    \bibliography{References}
}

\clearpage
\input{_Appendix}

\end{document}

%% file: 0_abstract.tex
\begin{abstract}

Existing adaptation techniques typically require architectural modifications or added parameters, leading to high computational costs and complexity.
We introduce Attention Projection Layer Adaptation (\apla), a simple approach to adapt vision transformers (ViTs) without altering the architecture or adding parameters.
Through a systematic analysis, we find that the layer immediately after the attention mechanism is crucial for adaptation.
By updating only this projection layer, or even just a random subset of this layer's weights, \apla achieves state-of-the-art performance while reducing GPU memory usage by up to \ViTgiantMemReduction and training time by up to \ViTgiantTimeReduction, with no extra cost at inference.
Across \TotalDatasets datasets covering a variety of tasks including scene classification, medical imaging, satellite imaging, and fine-grained classification, \apla consistently outperforms \NumAdaptationMethods other leading adaptation methods, including full fine-tuning, on classification, segmentation, and detection tasks. 
The code is available at 
\href{https://github.com/MoeinSorkhei/APLA}{
\texttt{\textcolor{pinkcolor}{github.com/MoeinSorkhei/APLA}}}.
\end{abstract}

%% file: 1_introduction.tex
\section{Introduction}
\label{intro}

\begin{figure}[t]
\centering
    \includegraphics[width=1.0\columnwidth]{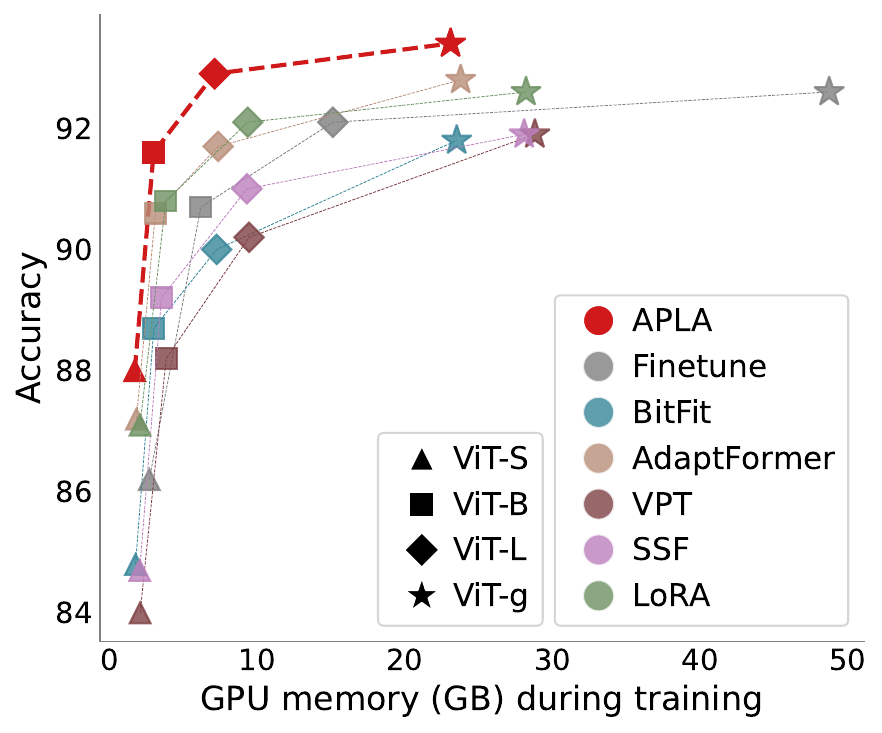}
\caption{
\emph{\apla achieves state-of-the-art for ViT adaption.} It yields better performance for a given GPU memory budget during training compared to full fine-tuning and leading adaptation methods. Similar savings are observed at inference 
(see Appendix \ref{apx:more_exp_results}).
}
\label{fig:figure_1_memory}
\vspace{-4mm}
\end{figure}

The primary objective of model adaptation is to enable models to generalize to new tasks with minimal data and computational cost.
The most successful approaches accomplish this by injecting new parameters or layers into frozen foundation models \cite{lian2022scaling, chen2022adaptformer, jia2022visual, hu2021lora}. 
This process often requires complex heuristics -- such as gradient sensitivity analyses \cite{he2023sensitivity, zhang2024gradient} and neural architecture searches \cite{zhang2022neural}—to determine the optimal locations to inject parameters.
Furthermore, the addition of new parameters can introduce significant overhead.
Aiming at better efficiency, a handful of methods attempt to adapt the existing structure of the model without adding parameters \cite{zhang2024gradient, zaken2021bitfit}, but these methods underperform compared to parameter-adding techniques.
This raises a critical question: is it possible to achieve competitive adaptation using only a model’s existing architecture? 
We propose that the answer is \emph{yes} -- and that the key is to better leverage the model’s inherent architecture.

We systematically investigate which existing components of a Vision Transformer (ViT) foundation model are most essential for adaptation in a departure from traditional parameter-adding approaches. 
Our analysis reveals that the projection layer immediately following the multi-head self-attention (MSA) mechanism plays a uniquely critical role. 
Then, inspired by low-rank approximation techniques \cite{hu2021lora, jie2023fact}, which demonstrate that updates to a full weight matrix can often be effectively represented with lower-dimensional matrices, we explore whether updating the entire projection layer is necessary. 
We find that \emph{modifying only a random subset of this layer’s parameters} is sufficient to maintain -- or even surpass -- performance, while further reducing computational costs. 
This result suggests that additional parameters used to learn a low-rank approximation of the updates may be unnecessary, opening the door for simpler, more efficient adaptation strategies.

In this work, we introduce Attention Projection Layer Adaptation (\apla), a novel state-of-the-art approach for efficient adaptation of ViTs that requires no additional parameters.
Our key contributions are as follows:
\begin{itemize}
    \item \textbf{Identification of a critical ViT component:} Through systematic experimentation, we identify the projection layer immediately following the attention mechanism as the most essential component for adaptation, offering a targeted approach to ViT tuning essential to \apla which can also improve other adaptation techniques.
    \item \textbf{Low-rank subset update for efficient adaptation:} Building on this insight, we introduce a low-rank adaptation technique that updates only a \textit{random subset} of the projection layer’s weights, achieving higher performance with even lower computational costs.
    \item \textbf{Simplified adaptation with no extra parameters:} Our method achieves SOTA results without introducing any new parameters, and eliminates the need for costly heuristics to determine where to inject new parameters or adapt existing ones.
    \item \textbf{Validated across scales and diverse applications:} We validate \apla on \TotalDatasets datasets across various tasks and model sizes, demonstrating its consistent superiority over \NumAdaptationMethods adaptation methods. \textit{In most cases, \apla performs better than full fine-tuning} while also achieving up to \ViTgiantMemReduction in GPU memory savings and a \ViTgiantTimeReduction reduction in training time.
\end{itemize}

\noindent
Together, these contributions establish \apla as a new standard for efficient and accessible ViT adaptation.

%% file: 2_related.tex
\section{Related Work}
\label{related}

Foundation models \cite{bommasani2021opportunities} have transformed computer vision, but as these models grow larger \cite{dehghani2023scaling, oquab2023dinov2, ilharco_gabriel_2021_5143773}, their fine-tuning requires high memory and computational resources. The traditional \textit{pretrain, then fine-tune} paradigm \cite{cui2018large, mustafa2021supervised, liu2021swin, zheng2021rethinking} has driven the field for years, but is becoming unfeasible for many applications due to these increasing costs. 
Recent advancements in model size \cite{dehghani2023scaling, oquab2023dinov2, ilharco_gabriel_2021_5143773} have only increased these challenges, making full fine-tuning unfeasible for many applications. 

In response, efficient adaptation methods have emerged, allowing practitioners with fewer resources to leverage large foundation models by introducing only a small set of new parameters, often called parameter-efficient fine-tuning (PEFT).
These methods reduce overhead, making large models more deployable in limited-resource settings.

\textit{Adapter-based methods} introduce compact, lightweight modules into specific layers, enabling task-specific adaptation by tuning only the adapter parameters while keeping the base model largely frozen. 
Originally developed for NLP, Adapters \cite{houlsby2019parameter} place bottleneck modules sequentially after each multi-head attention and MLP block \cite{vaswani2017attention}. 
AdaptFormer \cite{chen2022adaptformer} extends this for vision transformers, placing adapters in parallel with the MLP blocks rather than sequentially.
More recent methods refine adapter designs for greater efficiency. 
ARC \cite{dong2024efficient} uses a similar bottleneck operation but introduces parameter-sharing. 
SPT-Adapter \cite{he2023sensitivity} identifies and adapts only the most impactful layers based on gradient magnitudes. 
SSF \cite{lian2022scaling} appends learnable scaling and shifting transformations to modulate features after each ViT layer, while Consolidator \cite{hao2023consolidator} adds grouped connected layers that capture richer information through channel-wise input groups.
Adapter-based methods provide flexible, efficient model adaptation with fewer parameters than full fine-tuning. 
However, they increase inference costs, require careful initialization \cite{steitz2024adapters, houlsby2019parameter}, and their placement often relies on heuristics \cite{chen2022adaptformer} or gradient-based selection \cite{he2023sensitivity}, adding computational overhead and potentially leading to suboptimal configurations.

\textit{Low-rank-based methods} leverage the low-rank structure in adaptation updates, enabling efficient adaptation through low-rank matrices. 
LoRA \cite{hu2021lora} pioneered this approach by adding low-rank matrices alongside original weights in attention blocks. 
SPT-LoRA \cite{he2023sensitivity} builds on LoRA by selectively applying low-rank updates to layers with the largest gradient magnitudes. 
FacT \cite{jie2023fact} and RLRR \cite{dong2024low} decompose updates into factors, applying these across all ViT layers. 
While low-rank methods reduce adaptation costs, they insert additional parameters similarly to adapters.

\textit{Prompt-based} methods introduce learnable tokens to guide adaptation without modifying core model parameters. VPT \cite{jia2022visual} adds tokens to the input of each transformer block, and E$^2$VPT \cite{han20232vpt} incorporates auxiliary tokens into attention layers as well. Though prompt-based methods avoid changing internal parameters, they can increase inference costs due to the added tokens. NOAH \cite{zhang2022neural} combines prompts with adapters and LoRA modules, using neural architecture search to optimize placement.

\textit{Parameter-selective tuning} is an approach used by a handful of methods most closely related to \apla, that focus on adapting models by tuning only a subset of their existing parameters. 
GPS \cite{zhang2024gradient} selects parameters for tuning based on their gradient magnitudes, targeting the most error-inducing parameters during adaptation. 
BitFit \cite{zaken2021bitfit} takes a simpler approach, updating only the bias parameters. 
While these methods can be computationally efficient and easy to implement, they face challenges in identifying an optimal subset of parameters, which is reflected in their comparatively poor performance. 
\apla addresses this by identifying and targeting a critical layer for adaptation in ViTs, achieving state-of-the-art performance.

%% file: 3_methods.tex
\section{Methods}
\label{methods}

Inspired by methods that tune a subset of network weights and approaches that use low-rank updates, we ask, \enquote{\textit{Can we combine the strengths of both?}} 
To this end, we identify the most impactful ViT components for adaptation and propose a simple method that updates a low-rank subset of existing weights.

\subsection{Investigating adaptability of ViT components}

\begin{figure}[t]
\centering
\includegraphics[width=\columnwidth]{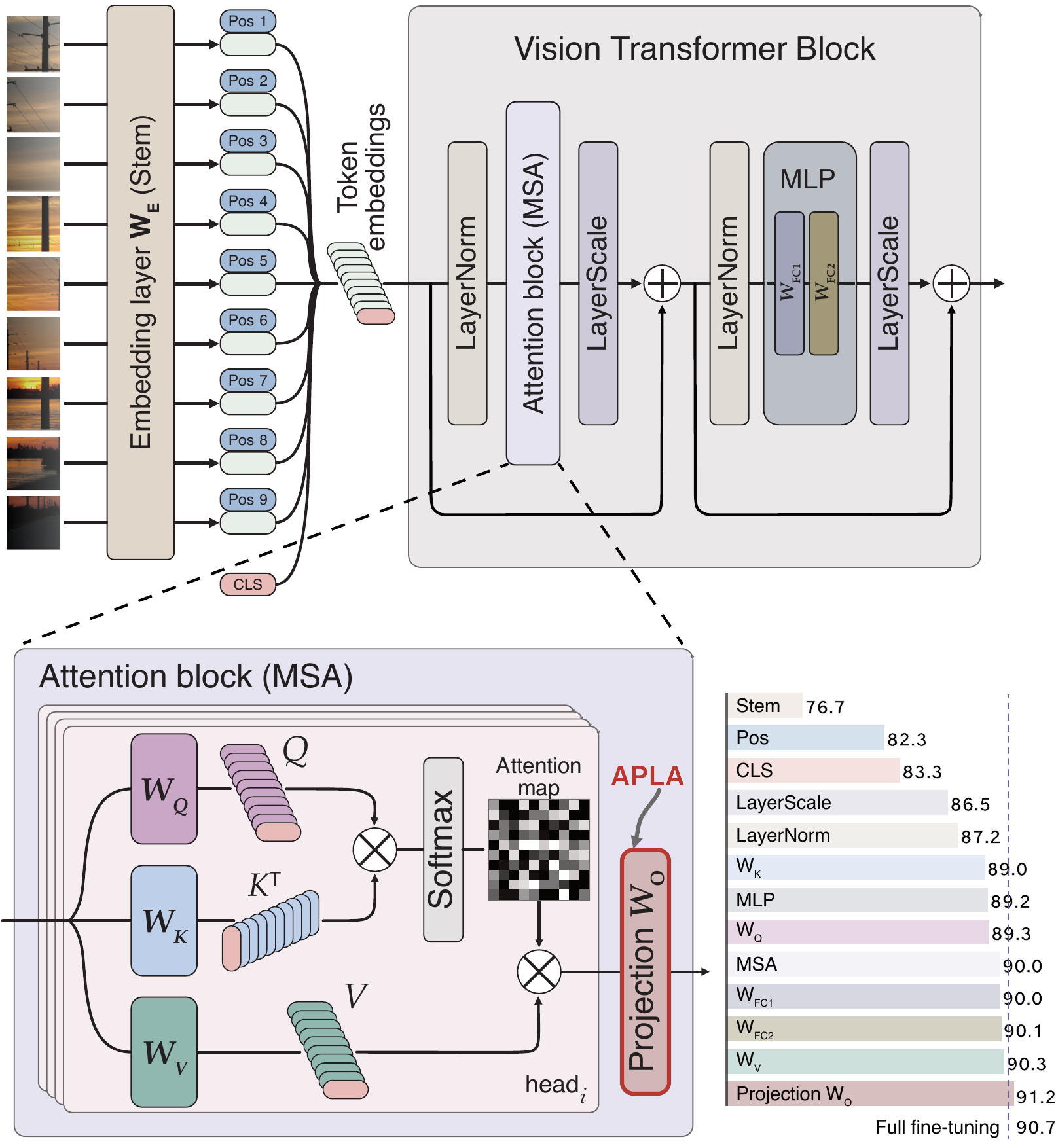}
\caption{
\emph{Investigating adaptation performance of individual ViT components.}
We evaluate the adaptation effectiveness of each ViT component in isolation across various downstream tasks, reporting the average performance. 
Results show that the attention output projection layer (\wo), located immediately after the attention mechanism, is the most effective for adaptation.
See Section~\ref{sec:choosing} and Table \ref{tab:which-component-to-tune} for detailed results.
}
\label{fig:figure_2-components-up-and-down}
\vspace{-4mm}
\end{figure}

A ViT is composed of multiple learnable components. 
To identify the most impactful ones for adapting the model to downstream tasks, we first review the different ViT components, grouped by their function (Figure \ref{fig:figure_2-components-up-and-down}).

Starting with an input image \(x\), a patchifying stem tiles and reshapes it into \(N\) flattened patches. 
Each patch undergoes a linear transformation in the embedding layer \(W_E\) with positional embeddings \(\text{Pos}_{n}, n \in \{1, \dots, N\}\) added to capture spatial information and a  classification token \cls appended to create the initial embeddings \(z_0\). 
These embeddings are passed through \(L\) transformer blocks, each containing LayerScale (LS) \cite{touvron2021going}, LayerNorm (LN) \cite{ba2016layer}, multi-head self-attention (MSA), and multi-layer perceptron (MLP) modules. 
The final representation is typically derived from the \cls token of the \(L\)th block, which is then processed by classification head \(W_{\text{pred}}\) to produce the prediction \(\hat{y}\).

In the MSA block, self-attention is computed for the input tokens using the learnable matrices \(W_{Q_i}\), \(W_{K_i}\), and \(W_{V_i}\), where \(i \in \{1, \dots, h\}\) corresponds to \(h\) parallel self-attention heads, allowing each head to learn distinct contextual relationships.
The self-attention output for each head is given by:
\begin{equation}    
    \text{head}_i = \text{softmax}\left( \frac{(z_{in} W_{Q_i}) (z_{in} W_{K_i})^T}{\sqrt{d_h}} \right) (z_{in} W_{V_i})   
    \label{eq:self_attentions}
\end{equation}
\noindent 
where \(d_h\) is the dimensionality of the Query, Key, and Value vectors for each self-attention head. 
The outputs from each \(\text{head}_i\) are concatenated, and a projection layer \( W_O \) re-weights the combined features to form the final output of the MSA block.
\begin{equation}
    z_{out} = [\text{head}_1 ; \text{head}_2 ; \dots ; \text{head}_h] W_{O}
    \label{eq:msa_projection}
\end{equation}
This output is then processed by an MLP block, consisting of two fully connected layers, \(W_{FC_1}\) and \(W_{FC_2}\), with a non-linearity in between. 

To identify the most essential component for adaptation, we conducted an empirical investigation by selectively tuning each of the components described above, one at a time (Figure \ref{fig:figure_2-components-up-and-down}) along with the final classification head \(W_{\text{pred}}\), while keeping the rest of the network frozen. 
We found that tuning only the projection layer \(W_O\)—positioned directly after the self-attention operation in the MSA block—yields the best performance, even surpassing full fine-tuning.
Details of this study are provided in Section~\ref{sec:choosing}.

\subsection{Low-rank adaptation through partial gradients}

Low-rank adaptation methods leverage the insight that the difference between initial and adapted values of a full-rank matrix can be closely approximated by a low-rank matrix
\[
W_{\text{approx}} \approx W_{\text{final}} - W_{\text{init}}, \quad \text{rank}(W_{\text{approx}}) \leq d
\]
where \(W_{\text{init}}\) and \(W_{\text{final}}\) are the layer's learnable matrix before and after adaptation, and \(d\) is the full rank.
Prior works (\eg \cite{hu2021lora, jie2023fact}) approximate this difference by adding low-rank matrices to ViT layers, with the rank \(r := \text{rank}(W_{\text{approx}})\) set as a hyperparameter.

In contrast, we propose a simpler low-rank adaptation by computing gradients on a \textit{randomly selected} subset of columns, which achieves substantial computational savings and retains the benefits of low-rank updates without adding parameters or altering the model’s architecture.

Specifically, given a parameter matrix 
\( W \in \mathbb{R}^{d \times d} \), 
we partition it into a trainable sub-matrix \( W_t \in \mathbb{R}^{d \times r} \) where gradients are computed during training, and a frozen sub-matrix \( W_f \in \mathbb{R}^{d \times (d-r)} \), which remains unchanged.
\begin{align}
    W_t &= W[i, j_m] \qquad m = 1, 2, \ldots, r
\end{align}
\noindent 
where the brackets denote indexing with \( \{j_1, j_2, \ldots, j_r\} \subseteq \{1, 2, \ldots, d\} \), representing randomly selected trainable column indices, where \( r\) controls the rank of the update, reaching full rank when \( r = d \).

\addtolength{\tabcolsep}{+3.5pt}
\begin{table}[t]  
    \centering
    \scriptsize
    \caption{
    \emph{The importance of ViT components for adaptation.}
    We evaluate how tuning each ViT component in isolation affects performance, while keeping the rest of the model frozen.
    We report classification performance,
    with the \textbf{best} and \underline{second best} results highlighted; this notation is used in subsequent tables.
    }
    \input{tables/components}
    \label{tab:which-component-to-tune}
    \vspace{-2mm}
\end{table}
\addtolength{\tabcolsep}{-3.5pt}

\subsection{Attention Projection Layer Adaptation (APLA)}

Foundation models already encode a rich set of features, and we hypothesize that adapting to new tasks can be achieved by selectively re-weighting these features to fit the target task.
Therefore, our approach to efficient model adaptation focuses on identifying impactful layers and computing partial gradients on a randomly selected subset of output features (matrix columns).

The projection layer $W_O$ is an ideal target for adaptation as it plays a central role in re-weighting features from the attention mechanism across all the heads. 

Therefore, we propose \textit{Attention Projection Layer Adaptation} (\apla), which tunes a randomly selected subset of columns in the \(W_O^l\) matrices in each transformer block, while keeping the rest of the ViT backbone frozen.
Specifically, we tune only a subset of column vectors of the \(W_O^l\) matrices and the final classification head \(W_{\text{pred}}\): 
\begin{equation}
    \mathcal{S}_{\text{APLA}} = \{W_O^1[i,j_m^1],W_O^2[i,j_m^2], {\dots}, W_O^L[i,j_m^L], {W_{\text{pred}}}\}
\end{equation}
\noindent 
For each transformer block \(l \leq L\), we independently sample a distinct subset of column indices \( \{j_1^l, j_2^l, \ldots, j_r^l\} \subseteq \{1, 2, \ldots, d\} \) given a global rank hyperparameter \( r \leq d \), chosen once at the beginning of training. 

APLA is easy to implement, computationally efficient, requires no new parameters, and introduces no additional inference latency, making it highly practical.

%% file: tables/components.tex
\begin{tabular}{@{} lccccc @{}}
\toprule

{} &  \birds & \cars & \aid & \isic & \average \\
\hline
\cls token                          &  83.9 & 89.9 & 92.9 & 66.4 & 83.3 \\
positional embeddings               &  85.1 & 89.6 & 92.9 & 61.4 & 82.3 \\
Embedding layer \(W_E\)              &  85.7 & 88.2 & 87.9 & 45.1 & 76.7 \\
\cdashlinelr{1-6}
LayerNorm                           &  83.7 & 91.5 & 94.6 & 79.0 & 87.2 \\
LayerScale                          &  85.5 & 91.2 & 93.9 & 75.4 & 86.5 \\
\cdashlinelr{1-6}
\wq weight matrix        &   85.5 & 91.8 & 94.6 & 85.1 & 89.3 \\
\wk weight matrix        &   85.8 & 91.8 & 94.5 & 83.8 & 89.0 \\
\wv weight matrix        &   85.8 & 93.2 & 95.3 & 86.9 & 90.3 \\
$\mathbf{W_O}$ \textbf{weight matrix}        &   \textbf{86.5} & \underline{94.0} & \textbf{96.0} & \textbf{88.2} & \textbf{91.2} \\
MSA block 
&   85.0 & 93.8 & 94.7 & 86.5 & 90.0 \\
\cdashlinelr{1-6}
\wfcone weight matrix       &   84.7 & 93.5 & 95.0 & 86.9 & 90.0 \\
\wfctwo weight matrix       &   84.6 & 93.4 & 94.7 & \underline{87.7} & 90.1 \\
MLP block 
&   82.4 & 93.6 & 94.3 & 86.4 & 89.2 \\
\cdashlinelr{1-6}
Full Finetuning      &   \underline{85.2} & \textbf{94.4} & \underline{95.4} & \underline{87.7} & \underline{90.7} \\
\bottomrule
\end{tabular}

%% file: 4_experimental_setup.tex
\section{Experimental Setup}
\label{sec:experimental_setup}

We benchmark APLA against \competeAgainst adaptation methods on \TotalDatasets datasets across \numberOfTasks tasks, using \foundationModels foundation models. 
For our main model types, we use Vision Transformers (ViTs) \cite{dosovitskiy2020image} and Swin Transformers \cite{liu2021swin} at varying capacities, unless stated otherwise. 
Below, we provide an overview of our experimental setup. 
Additional details are available in Appendix \ref{apx:more_exp_details}.

\vspace{-3mm}
\paragraph{Adaptation methods}
We evaluate \apla against various adaptation methods, beginning with traditional approaches: full fine-tuning (\finetune) and training only an appended linear layer (\linear). 
We further compare against \mlpK and \partialadaptK. 
In \mlpK, a $k$-layer MLP is appended to the model, and only this block is trained, while \partialadaptK tunes the last $k$ blocks of the model.
We set $k=3$ for \mlpK and $k=1$ for \partialadaptK, in line with \cite{jia2022visual}.
To benchmark efficient adaptation, we compare against \NumAdaptationMethods recent methods:
\bias \cite{zaken2021bitfit} ,
\adapter \cite{houlsby2019parameter}, 
\adaptformer \cite{chen2022adaptformer}, 
\vpt-shallow and \vpt-deep \cite{jia2022visual}, 
\etwovpt \cite{han20232vpt}, 
\ssf \cite{lian2022scaling}, 
\lora \cite{hu2021lora}, 
\spt-adapter and \spt-LoRa \cite{he2023sensitivity},
\noah \cite{zhang2022neural},
\fact-TK and \fact-TT \cite{jie2023fact}, 
\consolidator \cite{hao2023consolidator}
\arc \cite{dong2024efficient}, 
\gps \cite{zhang2024gradient}
and \rlrr \cite{dong2024low}.

\addtolength{\tabcolsep}{-5.5pt} 
\begin{table}[t]
\centering
    \caption{
    \emph{Comparing parameter-selection strategies for \apla.}
    }
    \centering
    \scriptsize
    \input{tables/selection}
    \label{tab:row-selection}
    \vspace{-2mm}
\end{table}
\addtolength{\tabcolsep}{+5.5pt}

\vspace{-3mm}
\paragraph{Datasets and tasks}
We benchmark \apla across \TotalDatasets datasets, covering a diverse set of object categories and tasks.
Starting with 21 generic image classification tasks, we cover superordinate object recognition, fine-grained classification, scene recognition, satellite imagery, and medical image analysis using the following datasets:
CUB-200-2011 \cite{wah2011caltech}, 
NABirds \cite{van2015building}, 
Birdsnap \cite{berg2014birdsnap}, 
Stanford Dogs \cite{khosla2011novel}, 
StanfordCars \cite{krause20133d}, 
Aircraft \cite{maji2013fine},  
Caltech-256 \cite{griffin2007caltech}, 
Caltech-101 \cite{fei2006one}, 
CIFAR-100 and CIFAR-10 \cite{krizhevsky2009learning}, 
Oxford-III Pet \cite{parkhi2012cats}, 
DTD \cite{cimpoi2014describing},  
MIT Indoor \cite{quattoni2009recognizing}, 
SUN397 \cite{xiao2010sun}), 
AID \cite{xia2017aid}, 
RSSCN7 \cite{cheng2017remote}, 
ISIC2019 \cite{tschandl2018ham10000, codella2018skin, combalia2019bcn20000}, 
APTOS2019 \cite{aptos2019-blindness-detection}, 
DDSM \cite{lee2017curated}, 
Colorectal \cite{kather2016multi}, 
and Pneumonia \cite{kermany2018identifying}.

\addtolength{\tabcolsep}{-4.15pt}
\begin{table*}[t]
    \centering
    \scriptsize
    \caption{
    \emph{Main results comparing adaptation methods on image classification} for ViT-B \cite{dosovitskiy2020image} pre-trained with \dinovtwo \cite{oquab2023dinov2}.
    The \textbf{best} and \underline{second} best results are highlighted for each task. 
    }
    \input{tables/general}
    \label{tab:general-classification}
\end{table*}
\addtolength{\tabcolsep}{+4.15pt}

For semantic \& instance segmentation and object detection we use \ade \cite{zhou2019semantic, zhou2017scene} and \coco \cite{lin2014microsoft}.

To test in low-data settings, we use \vtabOneK \cite{zhai2019large}, a collection of 19 classification tasks with only 1,000 training examples each—representing challenging yet realistic scenarios for model adaptation. 
For out-of-distribution (OOD) evaluation, we use ImageNet, ImageNet-A \cite{hendrycks2021natural}, ImageNet-C \cite{hendrycks2019benchmarking}, and ImageNet-R \cite{hendrycks2021many}, which introduce various domain shifts, including corruptions, perturbations, and adversarial examples.
We use the standard evaluation metric for each dataset.

\vspace{-3mm}
\paragraph{Foundation models}
To assess the impact of the foundation model type and pretraining strategy, we compare models trained on \imagenettwentyonek \cite{deng2009imagenet} using supervised learning, \openclip \cite{ilharco_gabriel_2021_5143773} trained with semi-supervision, and \dinovtwo \cite{oquab2023dinov2}, pre-trained using self-supervision.
We utilize \dinovtwo unless stated otherwise.

\vspace{-3mm}
\paragraph{Implementation details}
To ensure fair comparison, we closely follow the implementations in \cite{jia2022visual, dong2024efficient, dong2024low, jie2023fact, han20232vpt, chen2022adaptformer, hu2021lora}. 
For each dataset, we use the official protocol and standard train/val/test splits when available \cite{zhai2019large} or the splits provided by \cite{jia2022visual}. 
Models are trained with AdamW \cite{adamw} for 100 epochs, using a cosine decay learning rate schedule with a 10-epoch warm-up. 
Hyperparameters are selected via grid search on the validation set.

\apla, like other low-rank methods (\eg \cite{jie2023fact, hu2021lora}), requires selecting the value $r$, which in our case represents the number of columns of the weight matrix sampled for adaptation. 
We search over \( r \in \{8, 16, 128, 512, 768\} \) to find the optimal value.
Additional details regarding the experimental setup can be found in Appendix~\ref{apx:more_exp_details}.

%% file: tables/selection.tex
\begin{tabular}{l@{}ccccccc @{}}
\toprule
{} & 
\adjustbox{angle={35}}{NABirds} & 
\adjustbox{angle={35}}{SUN397} & 
\adjustbox{angle={35}}{Cal-256} & 
\adjustbox{angle={35}}{Cal-101} & 
\adjustbox{angle={35}}{Oxf.-Pet} & 
\adjustbox{angle={35}}{DDSM} & 
\adjustbox{angle={35}}{Average}
\\
\midrule
Largest gradients & 
\textbf{88.1} & 78.3 & 95.3 & 97.4 & 95.9 & \textbf{97.2} & 92.0 \\
Largest activations & 
87.8 & \underline{78.5} & \underline{95.6} & 97.7 & \textbf{96.1} & 96.7 & \underline{92.1} \\
Largest weight magnitude & 
87.8 & \textbf{78.6} & 95.5 & \underline{97.8} & \underline{96.0} & 97.0 & \underline{92.1} 
\\
\textbf{Random (\apla)} & 
\underline{88.0} & 78.3 & \textbf{95.7} & \textbf{98.0} & \textbf{96.1} & \textbf{97.2} & \textbf{92.2} \\
\bottomrule
\end{tabular}

%% file: tables/general.tex
\begin{tabular}{@{} l@{\hspace{1mm}}ccccccccccccccccccccccccccccccc@{\hspace{1mm}}c @{}}
\toprule
{}  & 
\multicolumn{7}{c}{\textbf{Fine-grained}} &&
\multicolumn{7}{c}{\textbf{General}} && 
\multicolumn{3}{c}{\textbf{Scene}} &&
\multicolumn{3}{c}{\textbf{Satellite}} && 
\multicolumn{6}{c}{\textbf{Medical}}  && 
\\
\cline{2-8}\cline{10-16}\cline{18-20}\cline{22-24}\cline{26-31}
\\
{} & 
\adjustbox{angle=\rotateN}{CUB-200-2011} & 
\adjustbox{angle=\rotateN}{NABirds} & 
\adjustbox{angle=\rotateN}{Birdsnap} & 
\adjustbox{angle=\rotateN}{StanfordDogs} & 
\adjustbox{angle=\rotateN}{StanfordCars} & 
\adjustbox{angle=\rotateN}{Aircraft} & 
\adjustbox{angle=\rotateN}{Average} &&

\adjustbox{angle=\rotateN}{Caltech-256} & 
\adjustbox{angle=\rotateN}{Caltech-101} & 
\adjustbox{angle=\rotateN}{CIFAR-100} & 
\adjustbox{angle=\rotateN}{CIFAR-10} & 
\adjustbox{angle=\rotateN}{Oxford-IIIT Pet} & 
\adjustbox{angle=\rotateN}{DTD} & 
\adjustbox{angle=\rotateN}{Average} &&

\adjustbox{angle=\rotateN}{MIT-Indoor} & 
\adjustbox{angle=\rotateN}{SUN397} & 
\adjustbox{angle=\rotateN}{Average} &&

\adjustbox{angle=\rotateN}{AID} & 
\adjustbox{angle=\rotateN}{RSSCN7} & 
\adjustbox{angle=\rotateN}{Average} && 

\adjustbox{angle=\rotateN}{ISIC2019} & 
\adjustbox{angle=\rotateN}{APTOS2019} & 
\adjustbox{angle=\rotateN}{DDSM} & 
\adjustbox{angle=\rotateN}{Coloretal} & 
\adjustbox{angle=\rotateN}{Pneumonia} & 
\adjustbox{angle=\rotateN}{Average} &&

\adjustbox{angle=\rotateN}{Total Average} 
\\
\midrule
\finetune & 
88.9 & 85.2 & 78.7 & 86.0 & \underline{94.4} & \textbf{87.5} & 86.8 &&
93.9 & 97.3 & 92.4 & 98.7 & 94.7 & 81.9 & 93.1 &&
87.8 & 75.6 & 81.7 &&
95.4 & 73.3 & 84.4 &&
\underline{87.7} & 90.8 & 95.5 & 97.8 & \underline{99.4} & 94.2 &&
89.7 
\\
\midrule
\linear & 
89.1 & 86.6 & 79.4 & 87.6 & 88.4 & 76.5 & 84.6 &&
94.9 & 97.0 & 88.9 & 98.0 & 95.8 & 81.1 & 92.6 &&
88.9 & 76.4 & 82.7 &&
91.2 & 77.1 & 84.2 && 
55.3 & 90.4 & 89.4 & 94.0 & 97.9 & 85.4 &&
86.9
\\
\mlp & 
89.1 & 86.4 & 78.9 & 87.8 & 88.3 & 77.7 & 84.7 && 
94.4 & 97.5 & 89.2 & 98.4 & \underline{96.0} & 80.9 & 92.7 &&
88.6 & 76.2 & 82.4 &&
91.6 & 76.4 & 84.0 && 
71.9 & 90.7 & 93.4 & 95.8 & 97.9 & 89.9 &&
88.0
\\
\partialadapt & 
88.8 & 86.5 & 78.6 & 87.4 & 88.1 & 76.6 & 84.3 &&
94.9 & 96.9 & 89.0 & 98.0 & \underline{96.0} & 80.6 & 92.6 && 
88.5 & 76.3 & 82.4 && 
90.9 & 77.7 & 84.3 && 
56.1 & 90.6 & 89.5 & 94.2 & 97.6 & 85.6 && 
86.8
\\
\midrule
\bias & 
89.4 & 87.9 & 80.7 & 87.8 & 92.5 & 83.2 & 86.9 && 
95.2 & 97.6 & 93.1 & \textbf{99.3} & 95.7 & 82.2 & 93.9 && 
89.3 & 77.7 & 83.5 && 
95.2 & 83.5 & 89.4 && 
79.0 & 90.1 & 96.4 & 97.2 & 99.1 & 92.4 && 
90.1 
\\
\adapter & 
\underline{89.6} & \textbf{88.4} & 80.0 & 87.8 & 93.5 & 86.1 & \underline{87.6} && 
95.0 & 97.6 & \underline{93.5} & \textbf{99.3} & 95.9 & 81.8 & 93.9 && 
89.5 & 77.9 & 83.7 && 
95.0 & 84.2 & 89.6 && 
84.3 & 89.6 & 96.1 & \underline{98.0} & 98.5 & 93.3 && 
90.6 
\\
\adaptformer & \textbf{89.7} & \textbf{88.4} & 80.5 & \underline{88.1} & 93.1 & 85.4 & 87.5 && 
95.6 & \textbf{98.1} & 93.2 & \textbf{99.3} & 95.9 & \underline{82.8} & \underline{94.2} && 
89.9 & \underline{78.1} & \underline{84.0} && 
95.4 & \underline{85.3} & \underline{90.4} && 
85.6 & 90.8 & \textbf{97.3} & 97.4 & 98.6 & 93.9 && 
\underline{90.9} 
\\
\vptshallow & 
88.8 & 86.7 & 79.1 & 87.4 & 90.6 & 73.2 & 84.3 && 
95.1 & 97.2 & 92.4 & 99.1 & \underline{96.0} & 80.4 & 93.4 && 
89.4 & 76.6 & 83.0 && 
91.6 & 70.4 & 81.0 && 
76.5 & 89.3 & 96.2 & 96.4 & 98.7 & 91.4 && 
88.1 
\\
\vptdeep & 
89.1 & 87.3 & 79.9 & 87.2 & 91.5 & 81.7 & 86.1 && 
95.3 & 97.8 & 92.7 & 99.1 & 95.7 & 80.5 & 93.5 && 
\underline{90.0} & 77.0 & 83.5 && 
94.4 & 78.0 & 86.2 && 
79.6 & 91.0 & 96.2 & 97.6 & 98.8 & 92.6 && 
89.5 
\\
\etwovpt & 
88.3 & 86.6 & 79.7 & 87.4 & 91.2 & 81.0 & 85.7 && 
94.5 & 96.9 & 92.7 & \underline{99.2} & 95.4 & 79.6 & 93.1 && 
88.7 & 76.3 & 82.5 && 
93.7 & 72.7 & 83.2 && 
80.9 & 90.6 & 96.2 & 96.8 & 98.6 & 92.6 && 
88.9 
\\
\ssf & 
89.4 & \underline{88.1} & 80.5 & 87.7 & 92.7 & 83.7 & 87.0 && 
95.3 & 97.8 & 93.2 & \underline{99.2} & 95.7 & 82.1 & 93.9 && 
88.9 & 77.4 & 83.2 && 
95.3 & 82.6 & 89.0 && 
80.7 & 90.5 & 96.4 & 97.2 & 99.0 & 92.8 && 
90.2 
\\
\lora & 
88.7 & 87.5 & 79.3 & 86.3 & 93.4 & 86.4 & 86.9 && 
94.3 & 97.1 & 93.0 & 99.0 & 94.0 & 80.4 & 93.0 && 
88.5 & 76.4 & 82.5 && 
95.4 & 81.4 & 88.4 && 
86.5 & \underline{91.1} & 95.1 & 97.4 & 98.9 & 93.8 && 
90.0 
\\
\sptadapter & 
89.4 & \underline{88.1} & 80.6 & 87.7 & 93.1 & 86.2 & 87.5 && 
\textbf{95.8} & 97.5 & 93.1 & \underline{99.2} & 95.8 & 82.7 & 94.0 && 
89.5 & \underline{78.1} & 83.8 && 
\underline{95.6} & 84.7 & 90.2 && 
82.1 & 90.4 & 96.1 & 97.2 & 99.0 & 93.0 && 
90.6 
\\
\sptlora & 
89.2 & 87.9 & 80.6 & 87.5 & 92.8 & 86.3 & 87.4 && 
\textbf{95.8} & 97.7 & 92.6 & \underline{99.2} & 95.7 & 82.3 & 93.9 && 
89.9 & 77.7 & 83.8 && 
95.4 & 84.2 & 89.8 && 
82.2 & 89.3 & 96.2 & 97.4 & 99.1 & 92.8 && 
90.4 
\\
\facttk & 
88.8 & 87.8 & 80.5 & 87.5 & 93.0 & 85.4 & 87.2 && 
95.3 & 97.6 & 92.8 & \underline{99.2} & 95.5 & 81.5 & 93.7 && 
88.9 & 77.4 & 83.2 && 
95.4 & 79.9 & 87.7 && 
85.1 & 91.5 & 96.3 & 97.2 & 97.9 & 93.6 && 
90.2 
\\
\facttt & 
88.8 & 87.6 & 79.7 & 87.1 & 92.9 & 84.3 & 86.7 && 
94.9 & 97.5 & 92.4 & \underline{99.2} & 95.5 & 81.7 & 93.5 && 
89.4 & 77.1 & 83.3 && 
94.5 & 80.6 & 87.6 && 
81.5 & 90.9 & 97.0 & 97.4 & 98.6 & 93.1 && 
89.9 
\\
\consolidator & 
\textbf{89.7} & 87.4 & \underline{81.5} & 87.1 & 93.0 & 83.4 & 87.0 && 
94.5 & 97.3 & 92.7 & 99.0 & 95.8 & 81.8 & 93.5 &&
89.5 & 77.0 & 83.3 &&  
94.6 & 78.0 & 86.3 &&  
81.9 & \underline{91.1} & 96.9 & 96.2 & \underline{99.4} & 93.1 && 
89.9
\\
\arc & 
89.4 & 88.2 & 80.7 & \underline{88.1} & 92.6 & 84.1 & 87.2 && 
\textbf{95.8} & 97.8 & 92.9 & \underline{99.2} & 96.0 & 82.8 & 94.1 && 
89.8 & 78.2 & 84.0 && 
95.6 & \textbf{86.2} & 90.9 && 
82.5 & 90.7 & \textbf{97.3} & 97.0 & 98.8 & 93.3 && 
90.7 
\\
\gps & 
89.1 & 86.4 & 80.7 & 86.6 & \textbf{94.7} & 85.5 & 87.2 &&
94.8 & 97.6 & \textbf{94.0} & \textbf{99.3} & 94.4 & 77.9 & 93.0 &&
88.4 & 76.9 & 82.7 && 
94.9 & 61.6 & 78.3 && 
87.7 & 90.8 & 96.7 & 97.4 & 99.3 & \underline{94.4} &&
89.3
\\
\rlrr & 
88.9 & 87.9 & 80.8 & 87.6 & 92.4 & 83.7 & 86.9 && 
95.2 & 97.4 & 93.1 & \underline{99.2} & 95.8 & 82.0 & 93.8 && 
89.5 & 77.6 & 83.6 && 
95.0 & 81.2 & 88.1 && 
81.7 & 90.4 & 96.3 & 96.6 & 98.6 & 92.7 && 
90.0 
\\
\midrule
\textbf{\apla} & 
\underline{89.6} & 88.0 & \textbf{81.9} & \textbf{88.5} & 94.0 & \underline{86.7} & \textbf{88.1} && 
\underline{95.7} & \underline{98.0} & 93.4 & \textbf{99.3} & \textbf{96.1} & \textbf{83.0} & \textbf{94.3} && 
\textbf{90.4} & \textbf{78.3} & \textbf{84.4} && 
\textbf{96.0} & \textbf{86.2} & \textbf{91.1} && 
\textbf{88.2} & \textbf{92.1} & \underline{97.2} & \textbf{98.4} & \textbf{99.5} & \textbf{95.1} && 
\textbf{91.5} 
\\

\bottomrule
\end{tabular}

%% file: 5_experiments.tex
\section{Experiments and Results}
\label{sec:experiments}

We begin by evaluating the choice of \apla’s core components, focusing first on identifying the most crucial layer for adaptation—the projection layer $W_O$—and how to select which parameters to tune within it. 
We then benchmark \apla against several other methods on standard classification, detection, and segmentation tasks, including low-data settings and out-of-distribution datasets. 
Additionally, we assess \apla across various model capacities, architectures, and foundation models. 
Finally, we evaluate its computational efficiency and explore how other adaptation techniques can benefit from our findings on the importance of the projection layer $W_O$.

\subsection{Choosing APLA's Components}
\label{sec:choosing}
\paragraph{Identifying which component to tune}

Previous work suggests that certain types of layers play a larger role in transfer learning \cite{yosinski2014transferable, sharif2014cnn, neyshabur2020being, matsoukas2022makes, konuklearning}. 
To identify the optimal components for \apla, we systematically investigate the effect of tuning different ViT components individually, keeping all other layers frozen.

Table \ref{tab:which-component-to-tune} and Figure \ref{fig:figure_2-components-up-and-down} present results across four mainstream datasets representing diverse domains, tasks, and data availabilities. 
We find that tuning the projection layer \wo yields the best performance, even outperforming full-network fine-tuning. 
The clear advantage of \wo over other ViT components leads us to adopt it as the default layer to tune in \apla.

\vspace{-3mm}
\paragraph{Identifying which parameters to tune}

Motivated by the effectiveness of low-rank adaptation methods, we explore strategies to economize \apla by selecting specific parameters in \wo for adaptation. 
We evaluate various strategies, including selecting columns with the largest gradients, activations, and weight magnitudes, as well as selecting the columns randomly.

We report our findings in Table \ref{tab:row-selection}.
Surprisingly, tuning a random subset of columns in \wo performs on par with more sophisticated selection methods. 
This suggests that, within the projection layer \wo, the specific choice of tunable columns is less critical. 
For \apla we choose the random selection approach because it may better utilize feature redundancy across attention heads and achieves a slight advantage in performance without added computational cost.

\addtolength{\tabcolsep}{-3.4pt}
\begin{table}[t]
    \centering
    \scriptsize
    \caption{
    \emph{Low-data settings.} 
    We benchmark on \vtabOneK which contains 19 low-data tasks grouped across three domains.
    }
    \input{tables/vtab_low_data_and_swin}
    \label{tab:vtab_low_data_and_swin}
\end{table}
\addtolength{\tabcolsep}{+3.4pt}

\subsection{Benchmarking APLA}
\label{sec:benchmark}

\paragraph{Mainstream classification tasks}

We benchmark \apla against other methods across 21 diverse image classification tasks, including superordinate classification, fine-grained classification, scene recognition, satellite imagery, and medical image analysis. 
Table \ref{tab:general-classification} presents the results.
On average, \apla outperforms all other methods, showing a 0.6\% improvement over the second-best method, \adaptformer. It ranks as the top or second-best method on 18 out of 21 datasets, consistently demonstrating strong performance across different classification tasks.

\vspace{-3mm}
\paragraph{APLA in low-data settings}

Foundation model adaptation is especially valuable in low-data scenarios, where reusing pretrained features is essential, as randomly initialized models tend to underperform \cite{dosovitskiy2020image, kolesnikov2020big, matsoukas2023pretrained}. 
We evaluate \apla in this regime using the \vtabOneK benchmark, which includes 19 datasets, each with only 1,000 samples. 
We use ViT-B and Swin-B pretrained on \imagenettwentyonek and report average performance across three domains: natural, specialized, and structured.
As shown in Table \ref{tab:vtab_low_data_and_swin}, \apla outperforms all other methods, achieving at least a 1\% improvement across these domains.
Visualizations \cite{van2008visualizing} of the \cls embeddings of different adaptation methods from these experiments are provided in Figure \ref{fig:tsne_extra} in the Appendix.

\addtolength{\tabcolsep}{+3.0pt}
\begin{table}[t]
    \centering
    \scriptsize
    \caption{
    \emph{OOD robustness.} 
    We assess robustness to OOD data by adapting on ImageNet-1K and testing on ImageNet-A, ImageNet-R, and ImageNet-C.
    }
    \input{tables/ood}
    \label{tab:ood}
\end{table}
\addtolength{\tabcolsep}{-3.0pt}

\addtolength{\tabcolsep}{-4.35pt}
\begin{table}[t]
    \centering
    \scriptsize
    \caption{
    \emph{Segmentation and detection.}
    Results for ADE20K semantic segmentation (SETR-PUP \cite{zheng2021rethinking} with a ViT-Large backbone) and COCO object detection \& instance segmentation (Mask R-CNN \cite{he2017mask} with a Swin-Tiny backbone).
    }
    \input{tables/other_tasks}
    \label{tab:other-tasks}
\end{table}
\addtolength{\tabcolsep}{+4.35pt}

\vspace{-3mm}
\paragraph{Out-of-distribution robustness}

While \apla has shown strong performance across various settings, its robustness under domain shifts and adversarial examples remains to be assessed. 
Using a foundation model pre-trained on ImageNet-21K, tuned on ImageNet-1K with various adaptation methods, we evaluate on ImageNet-A \cite{hendrycks2021natural}, ImageNet-R \cite{hendrycks2021many}, and ImageNet-C \cite{hendrycks2019benchmarking}. 
As shown in Table \ref{tab:ood}, \apla outperforms other methods overall, achieving an 8.6\% improvement in mean corruption error (mCE) on ImageNet-C. 
Notably, \apla and most other adaptation methods outperform full fine-tuning across all OOD datasets, highlighting the potential of efficient adaptation methods for OOD tasks.

\vspace{-3mm}
\paragraph{Segmentation \& Detection Tasks}

We evaluate \apla on semantic segmentation, object detection, and instance segmentation. 
For semantic segmentation, we use SETR-PUP \cite{zheng2021rethinking} with a ViT-Large backbone pre-trained on \imagenettwentyonek, reporting mean Intersection over Union (mIoU) for single-scale (SS) and multi-scale (MS) evaluations on \ade, as in \cite{jia2022visual, he2023sensitivity}. 
For object detection and instance segmentation, we use Mask R-CNN \cite{he2017mask} with a Swin-Tiny backbone pre-trained on \imagenetOneK, following \cite{lian2022scaling, liu2021swin}, and report mean Average Precision (AP) for bounding box ($\mathrm{AP}^{\mathrm{bb}}$) and mask ($\mathrm{AP}^{\mathrm{m}}$) predictions on \coco \cite{lin2014microsoft}. 
Additional details are in Appendix \ref{apx:more_exp_details}. 
As shown in Table \ref{tab:other-tasks}, \apla surpasses all other adaptation methods, with particularly strong results for semantic segmentation.

\addtolength{\tabcolsep}{-4.4pt}
\begin{table}[t]
    \centering
    \scriptsize
    \caption{
    \emph{Impact of pre-training strategy.} ViT-B pre-trained with \imagenettwentyonek and \openclip, then adapted to various tasks.
    }
    \input{tables/init}
    \label{tab:init}
\end{table}
\addtolength{\tabcolsep}{+4.4pt}

\vspace{-3mm}
\paragraph{Different foundation model types and capacities.}

We evaluate \apla’s versatility across foundation model training strategies, capacities, and architectures, using both supervised \imagenettwentyonek and semi-supervised \openclip ViT-B models, as well as Swin transformers \cite{liu2021swin}. 
To assess scalability, we test ViT models of varying sizes (ViT-S, ViT-B, ViT-L, and ViT-g).

As shown in Tables \ref{tab:vtab_low_data_and_swin}, \ref{tab:init}, and \ref{tab:scale}, \apla consistently outperforms other methods regardless of pretraining, model size, or architecture, maintaining strong performance where other methods show inconsistent performance.

\vspace{-3mm}
\paragraph{Applying other adaptation methods on \wo}

In Section \ref{sec:choosing} we showed that \wo is the most impactful component to adapt—surpassing full fine-tuning—and in Section \ref{sec:benchmark} we show that, when targeted in \apla, it outperforms other adaptation methods.
We now explore what happens if other adaptation methods are applied to \wo.
Do they improve performance when targeted to this layer? 
Does the low-rank adaptation strategy we propose for \apla prevail against other adaptation methods that target the same layer?
Using ViT-B pretrained with \dinovtwo, we apply \lora \cite{hu2021lora}, \fact \cite{jie2023fact}, and \adaptformer \cite{chen2022adaptformer}
on \wo and compare on \birds, \cars, \aid, \isic, and \vtabOneK.

As shown in Table \ref{tab:why-apla}, applying other adaptation methods to \wo generally improves performance, solidifying the importance of the \wo layer. 
\textit{Critically, \apla still outperforms other leading approaches when they are applied to \wo, suggesting there is an advantage to our simple low-rank adaptation strategy using random partial gradients.} 

\addtolength{\tabcolsep}{+6.8pt}
\begin{table}[t]
    \centering
    \scriptsize
    \caption{
    \emph{Impact of model capacity.}
    Results are averaged across NABirds, StanfordCars, AID, and ISIC2019 using \dinovtwo pre-trained models.
    Detailed results can be found in Appendix  \ref{apx:model_scale}.
    }
    \input{tables/scale_summary}
    \label{tab:scale}
\end{table}
\addtolength{\tabcolsep}{-6.8pt}

\vspace{-3mm}
\paragraph{Computational cost}
We analyze the computational costs of adaptation methods by measuring GPU memory footprint, parameter count, and throughput during training and inference. 
Results for training are shown in Figure \ref{fig:computes}, with additional results and details in the Appendix \ref{apx:more_exp_results}.
\apla demonstrates significant efficiency improvements, reducing memory usage and boosting training throughput, with no extra inference cost. 
Figure \ref{fig:figure_1_memory} further illustrates that \apla’s memory savings increase with model size, even surpassing BitFit, which tunes only bias parameters. 
Figure \ref{fig:computes_extended} in the Appendix reports parameter count, showing that \apla appears more costly than many methods according to this metric. 
However, as noted by prior work \cite{dehghani2021efficiency, cai2020tinytl}, parameter count is misleading in assessing computational efficiency. 
In practice, \textit{\apla remains the most efficient method}, consistently outperforming others in GPU memory usage and throughput during both training and inference.

We also examine how increasing the rank $r$ affects computational requirements for \apla and other low-rank methods in Appendix \ref{apx:other_methods_wo}. 
In tasks with significant domain shifts, a higher $r$ may be needed to re-weight features and bridge the gap between pretraining and the target domain. 
By leveraging existing model parameters, \apla keeps memory and throughput costs nearly fixed, even at higher ranks, unlike other approaches that become increasingly expensive as $r$ increases. 

%% file: tables/vtab_low_data_and_swin.tex
\begin{tabular}{@{} lccccccccc @{}}
\toprule
{} &
\multicolumn{4}{c}{\textbf{ViT-B}} && 
\multicolumn{4}{c}{\textbf{Swin-B}} 
\\
\cline{2-5}\cline{7-10}
\\
{} &
\vtabnatural &  \vtabspecialized & \vtabstructured & \average &&
\vtabnatural &  \vtabspecialized & \vtabstructured & \average
\\
\midrule
\finetune & 
75.9 & 83.4 & 47.6 & 69.0 &&
79.1 & 86.2 & 59.7 & 75.0 
\\
\midrule
\linear & 
68.9 & 77.2 & 26.9 & 57.7 &&
73.5 & 80.8 & 33.5 & 62.6 
\\
\mlp & 
67.8 & 72.8 & 30.6 & 57.1 &&
73.6 & 75.2 & 35.7 & 61.5 
\\
\partialadapt & 
69.4 & 78.5 & 34.2 & 60.7 &&
73.1 & 81.7 & 35.0 & 63.3 
\\
\midrule
\bias & 
73.3 & 78.3 & 44.1 & 65.2 &&
74.2 & 80.1 & 42.4 & 65.6 
\\
\adapter & 
79.0 & 84.1 & 58.5 & 73.9 &&
81.7 & 87.3 & 61.2 & 76.7
\\
\adaptformer & 
80.6 & 85.4 & 58.8 & 74.9 &&
-- & -- & -- & -- 
\\
\vptshallow & 
76.8 & 79.7 & 47.0 & 67.8 &&
79.9 & 82.5 & 37.8 & 66.7
\\
\vptdeep & 
78.5 & 82.4 & 55.0 & 72.0 &&
76.8 & 84.5 & 53.4 & 71.6
\\
\etwovpt & 80.0 & 84.4 & 57.4 & 73.9 &&
\underline{83.3} & 85.0 & 57.4 & 75.2
\\
\ssf & 
81.6 & 86.6 & 59.0 & 75.7 &&
-- & -- & -- & -- 
\\
\lora & 
79.5 & 84.6 & 60.5 & 74.8 &&
81.7 & 87.2 & 60.1 & 76.3
\\
\sptadapter & 
82.0 & 85.8 & 61.4 & 76.4 &&
83.0 & 87.3 & \underline{62.1} & \underline{77.5}
\\
\sptlora & 
81.9 & 85.9 & 61.3 & 76.4 &&
83.1 & \underline{87.4} & 60.4 & 77.2
\\
\noah &	
80.2 & 84.9 & 61.3 & 75.5 &&
-- & -- & -- & -- 
\\
\consolidator & 
82.4 & 86.3 & 60.9 & 76.5 &&
-- & -- & -- & --
\\
\facttk & 
80.6 & 85.3 & 60.7 & 75.5 &&
-- & -- & -- & -- 
\\
\facttt & 
80.6 & 85.0 & 60.5 & 75.3 &&
83.1 & 86.9 & \underline{62.1} & 77.4
\\
\arc & 
81.8 & 87.0 & 61.4 & 76.7 &&
79.0 & 86.6 & 59.9 & 75.2
\\
\rlrr & 
\underline{83.7} & \underline{87.3} & 61.8 & \underline{77.6} &&
81.3 & 86.7 & 59.0 & 75.7
\\
\gps & 
\underline{83.7} & 86.8 & \underline{61.9} & 77.5 &&
-- & -- & -- & --
\\
\midrule
\textbf{\apla} & 
\textbf{84.6} & \textbf{88.5} & \textbf{62.7} & \textbf{78.6} &&
\textbf{84.4}	& \textbf{87.8}	& \textbf{65.9} & \textbf{79.4}
\\
\bottomrule
\end{tabular}

%% file: tables/ood.tex
\begin{tabular}{@{} l@{\hspace{-2mm}}cccc @{}}
\toprule
{} & 
ImageNet-1K &
ImageNet-A &
ImageNet-R &
ImageNet-C 
\\
{} & 
Acc. ($\uparrow$)& 
Acc. ($\uparrow$) & 
Acc. ($\uparrow$) & 
mCE ($\downarrow$)
\\

\midrule
\finetune & 
83.6 & 
34.5 & 
51.3 & 
46.5 \\
\hline
\linear & 
82.0 & 
33.9 & 
52.9 & 
46.9 \\
\hline
\adapter & 
82.7 & 
42.2 & 
54.1 & 
42.7 \\
\bias & 
82.7 & 
42.1 & 
55.9 & 
41.9 \\
\vptshallow & 
82.1 & 
30.9 & 
53.7 & 
46.9 
\\
\vptdeep & 
82.5 & 
39.1 & 
53.5 & 
43.1 
\\
\ssf & 
83.1 & 
45.9 & 
\underline{56.8} & 
\underline{41.5}
\\
\gps & 
\underline{83.9} & 
\underline{46.1} & 
\textbf{57.0} & 
42.0 
\\
\hline
\textbf{\apla} & 
\textbf{84.0} & 
\textbf{46.9} & 
55.5 & 
\textbf{32.9}
\\
\bottomrule
\end{tabular}

%% file: tables/other_tasks.tex
\begin{tabular}{@{} l@{\hspace{-1mm}}ccccccccc @{}}
\hline
\toprule
{} & 
\multicolumn{2}{c}{\textbf{ADE20K}} &&
\multicolumn{6}{c}{\textbf{COCO}}
\\
\cline{2-3}\cline{5-10}
\vspace{-1.5mm}
\\
{} & 
\adjustbox{angle=0}{\fontsize{4.5pt}{5pt}{mIoU (SS)}} & 
\adjustbox{angle=0}{\fontsize{4.5pt}{5pt}{mIoU (MS)}} && 
\adjustbox{angle=0}{\fontsize{5pt}{6pt}{$\mathrm{AP}^{\mathrm{bb}}$}} &
\adjustbox{angle=0}{\fontsize{5pt}{6pt}{$\mathrm{AP}^{\mathrm{bb}}_{50}$}} &
\adjustbox{angle=0}{\fontsize{5pt}{6pt}{$\mathrm{AP}^{\mathrm{bb}}_{75}$}} &
\adjustbox{angle=0}{\fontsize{5pt}{6pt}{$\mathrm{AP}^{\mathrm{m}}$}} &
\adjustbox{angle=0}{\fontsize{5pt}{6pt}{$\mathrm{AP}^{\mathrm{m}}_{50}$}} &
\adjustbox{angle=0}{\fontsize{5pt}{6pt}{$\mathrm{AP}^{\mathrm{m}}_{75}$}}
\\
\midrule
\bias & 
43.4 & 45.3 &&
33.7 & 57.8 & 35.0 & 32.7 & 54.7 & 33.9
\\
\vptdeep & 
42.1 & 44.1 &&
33.8 & 57.6 & 35.3 & 32.5 & 54.5 & 33.9
\\
\ssf &
\underline{45.6} & 47.4 &&
34.9 & 58.9 & 36.1 & 33.5 & 55.8 & 34.7

\\
\lora & 
43.9 & 45.9 &&
37.1 & 60.9 & 39.5 & 35.2 & 57.7 & 37.2
\\
\adapter & 
44.4 & 46.6 &&
\underline{37.6} & \underline{61.1} & \underline{40.2} & \underline{35.6} & \underline{58.2} & \underline{37.8}
\\
\adaptformer &
44.3 & 46.2 &&
35.1 & 59.1 &  36.9 &  33.8 &  56.0 & 35.6
\\
\sptadapter & 
45.2 & 47.2 &&
-- & -- & -- & -- & -- & --
\\
\sptlora & 
45.4 & \underline{47.5} &&
-- & -- & -- & -- & -- & --
\\
\midrule
\textbf{\apla} & 
\textbf{48.3} & \textbf{49.5} &&
\textbf{38.1} & \textbf{61.8} & \textbf{40.9} & \textbf{35.9} & \textbf{58.7} & \textbf{37.9}
\\
\bottomrule
\end{tabular}

%% file: tables/init.tex
\begin{tabular}{@{} lccccccccccc @{}}
\toprule
{}  & 
\multicolumn{5}{c}{\textbf{\imagenettwentyonek}} && 
\multicolumn{5}{c}{\textbf{\openclip}} 
\\
\cline{2-6}\cline{8-12}
\\
{} & 
\birds & 
\cars & 
\aid & 
\isic &  
\average &&

\birds & 
\cars & 
\aid & 
\isic &  
\average
\\
\midrule
\finetune & 82.7 & 84.5 & \underline{91.7} & \underline{84.0} & 85.7 &&
79.0 & 94.7 & 95.6 & 84.9 & \underline{88.6}
\\
\midrule
\linear & 75.9 & 51.3 & 81.0 & 51.2 &   64.9 && 
73.7 & 94.5 & 95.0 & 54.2 &  79.4
\\
\mlp & 77.3 & 54.9 & 81.2 & 61.7 &   68.8 && 73.6 & 93.8 & 95.0 & 68.7 &   82.8
\\
\partialadapt & 77.8 & 66.2 & 81.1 & 46.6 &   67.9 && 
73.8 & 94.4 & 95.2 & 54.9 &   79.6
\\
\midrule
\bias & 84.2 & 79.4 & 90.5 & 73.0 &   81.8 && 
79.2 & 95.0 & \underline{95.8} & 72.7 &   85.7 
\\
\adapter & 
84.3 & 68.6 & 90.0 & 80.6 &   80.9 && 
79.2 & 95.0 & 95.2 & \underline{83.9} &   88.3
\\
\adaptformer & 78.8 & 83.1 & 90.1 & 81.2 &   83.3 && 
\textbf{80.0} & 95.0 & 95.5 & 82.4 &   88.2 
\\
\vptshallow & 78.8 & 68.7 & 85.9 & 65.0 &   74.6 && 
73.8 & 94.7 & 95.1 & 54.6 &   79.6 
\\
\vptdeep & 84.2 & 83.6 & 89.0 & 74.8 &   82.9 && 
73.6 & 94.5 & 95.1 & 54.6 &   79.5 
\\
\etwovpt & 84.6 & 82.8 & 88.4 & 78.6 &   83.6 && 
77.7 & \underline{95.1} & \textbf{95.9} & 73.6 &   85.6 
\\
\ssf & \textbf{85.7} & 89.2 & 90.9 & 78.4 &   86.1 && 
\underline{79.9} & 94.9 & 95.7 & 76.1 &   86.7 
\\
\lora & 
\underline{85.6} & 83.2 & 91.0 & 83.5 & 85.8 &&
79.0 & 94.1 & 95.0 & 83.2 &   87.8 
\\
\sptadapter & 83.3 & 86.2 & 90.8 & 75.7 &   84.0 && 
76.0 & 94.8 & 95.4 & 76.4 &   85.7 
\\
\sptlora & 83.4 & 87.3 & 90.0 & 76.2 &   84.2 && 
75.9 & 94.9 & 95.1 & 77.4 &   85.8 
\\
\facttk & 80.3 & 84.0 & 91.1 & 79.3 &   83.7 && 
79.2 & 94.9 & 95.0 & 80.7 &   87.5 
\\
\facttt & 79.2 & 82.4 & 90.9 & 77.7 &   82.6 && 
78.5 & 94.9 & 95.4 & 77.2 &   86.5 
\\
\arc & \textbf{85.7} & 89.5 & 90.8 & 79.2 &  \underline{86.3} && 
79.5 & \underline{95.1} & 95.1 & 79.0 &   87.2 
\\
\rlrr & 85.3 & \underline{90.4} & 91.1 & 77.5 &   86.1 && 
80.0 & 94.9 & \underline{95.8} & 79.5 &   87.6 
\\
\midrule
\textbf{\apla} & 85.2 & \textbf{90.5} & \textbf{94.3} & \textbf{84.9} &   \textbf{88.7} && 
79.1 & \textbf{95.2} & \textbf{95.9} & \textbf{85.5} &   \textbf{88.9} 
\\
\bottomrule
\end{tabular}

%% file: tables/scale_summary.tex
\begin{tabular}{@{} lccccc @{}}
\toprule
{} & \vits & \vitb & \vitl & \vitg
\\
\midrule
\finetune & 
86.2 & 90.7 & \underline{92.1} & 92.6
\\
\midrule
\linear & 
76.1 & 80.4 & 83.6 & 85.5 
\\
\mlp & 
80.9 & 84.6 & 86.2 & 87.8 
\\
\partialadapt & 
75.8 & 80.4 & 83.6 & 85.2 
\\
\midrule
\bias & 
84.8 & 88.7 & 90.0 & 91.8 
\\
\adapter & 
\underline{87.9} & 90.3 & 91.9 & 92.3 
\\
\adaptformer & 
87.2 & 90.6 & 91.7 & \underline{92.8} 
\\
\vptshallow  & 
82.8 & 86.4 & 88.3 & 88.9 
\\
\vptdeep     & 
84.0 & 88.2 & 90.2 & 91.9 
\\
\etwovpt & 84.2 & 88.1 & 90.9 & 91.7 
\\
\ssf          & 
84.7 & 89.2 & 91.0 & 91.9 
\\
\lora         & 
87.1 & \underline{90.8} & \underline{92.1} & 92.6 
\\
\sptadapter  & 
87.2 & 89.7 & 90.0 & 89.5 
\\
\sptlora     & 
87.6 & 89.6 & 90.5 & 89.4 
\\
\facttk      & 
86.9 & 90.3 & 91.5 & 92.0 
\\
\facttt      & 
85.4 & 89.1 & 91.3 & 91.6 
\\
\arc          & 
86.0 & 89.7 & 91.1 & 91.9 
\\
\rlrr         & 
84.4 & 89.3 & 91.8 & 92.3 
\\
\midrule
\textbf{\apla}   & 
\textbf{88.0} & \textbf{91.6} & \textbf{92.9} & \textbf{93.4} 
\\
\bottomrule

\end{tabular}

%% file: 6_discussion.tex
\section{Discussion}
\label{discussion}

\addtolength{\tabcolsep}{-3.1pt} 
\begin{table}[t]
    \caption{
    \emph{Applying other adaptation methods on \wo}. 
    Competing methods are applied to \wo,  isolating the effect of \apla's low-rank adaptation strategy.
    ``Natural'', ``Specialized'', and ``Structured'' indicate average performance across respective VTAB groups.
    Notably, \lora and \adaptformer would be improved if they were placed at \wo rather than their default locations.
    }
    \centering
    \scriptsize
    \input{tables/proj_only}
    \label{tab:why-apla}
    \vspace{-2mm}
\end{table}
\addtolength{\tabcolsep}{+3.1pt}

\textit{\apla establishes a new state-of-the-art for efficient model adaptation} across a wide range of classification, segmentation, and detection tasks, showing resilience in low-data and out-of-distribution scenarios.
It achieved top performance across model types, capacities, and pretraining methods -- a level of versatility that no other adaptation method maintained across such varied conditions. 
Moreover, apla excels in computational efficiency, significantly reducing memory and processing requirements.

 \textit{\apla offers several practical advantages} in addition to its exceptional performance. 
 It is simple to implement, requiring no architectural changes or added parameters which may be sensitive to initialization.
 It eliminates the need to search for which parameters to update.
 \apla's simplicity makes it easy to work with. 
 \apla also supports flexible layer adaptation, allowing partial or full layer updates depending on the need, all with minimal computational overhead. 
 Finally, as demonstrated in Section \ref{sec:benchmark}, \apla's core insights can be used to enhance other adaptation methods, \eg~, applying \textsc{AdaptFormer} solely on \wo gives a $1$\% boost in performance.

\textit{What makes \apla so effective?} Although we don’t have a definitive answer, we offer two possible explanations.
One is the targeting of \wo.
Foundation models encode a rich set of features robust to various tasks. 
However, each task benefits from a unique composition of these features, making feature re-weighting essential. 
This is precisely the role of \wo, which re-weights the contribution of features across all attention heads.
Figure \ref{fig:figure_2-components-up-and-down} reveals that other top-performing ViT components serve similar functions: $W_{V}$ re-weights the attention output within each head, while \wo operates across all heads. 
Given this, one might expect \(W_{FC_1}\) and \(W_{FC_2}\) in the MLP block to play a more critical role, but they are positioned further downstream, with \wo and normalization layers modifying the features before they reach the MLP block.

A second explanation for \apla's effectiveness lies in the simplicity of its low-rank adaptation using randomly selected gradient updates.
Other approaches to use heuristics to select parameters, \eg~based on large weights, activations, or gradients may be suboptimal for foundation models, which are highly over-parameterized and exhibit feature redundancy. 
Selection based on large gradients or weights may not capture the most relevant features, could bias adaptation toward redundant or overly specific features, and lead to overfitting. 
By re-weighting a broader range of features, random selection makes \apla equally or even more effective in contexts of high feature redundancy, as shown in Table \ref{tab:row-selection}.

\textit{Why does \apla often outperform full fine-tuning}?
We speculate that, in low-data settings, full fine-tuning may distort pre-trained features to fit the limited training set, leading to overfitting--an issue less pronounced with partial tuning of \wo.

\begin{figure}[t]
\centering
\includegraphics[width=1.0\columnwidth]{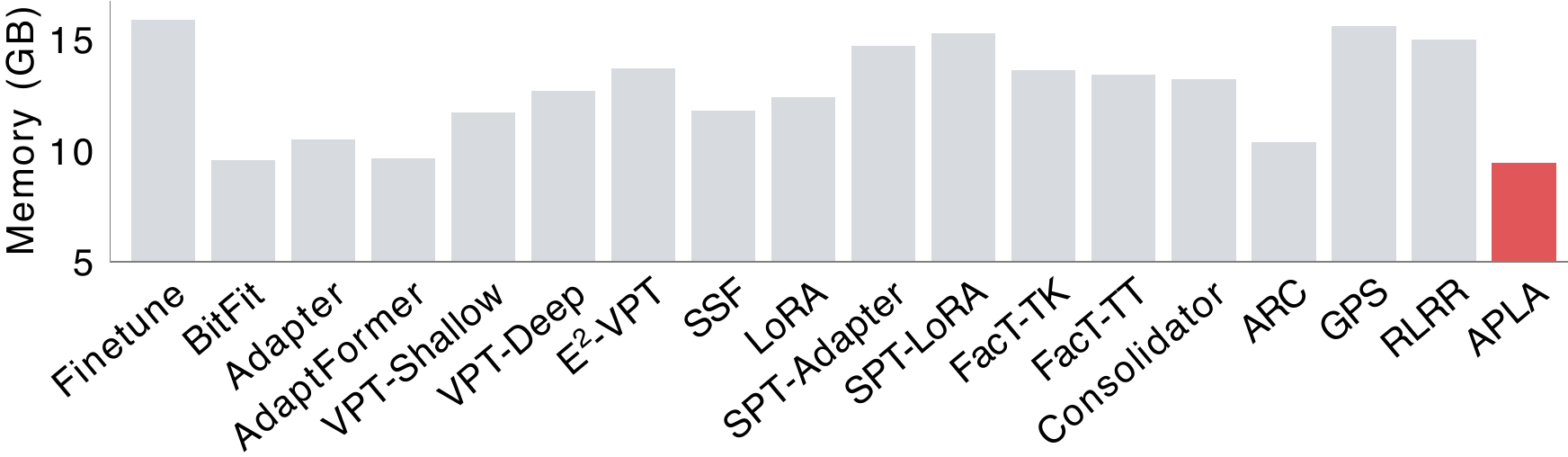}
\caption{
\emph{Computational cost.}
We report the memory footprint of various adaptation methods during training with a batch size of 64.
Results for parameter count, memory, and throughput during training and inference are provided in Appendix  \ref{apx:more_exp_results_computes}.
}
\vspace{-2mm}
\label{fig:computes}
\end{figure}

\vspace{-3mm}
\paragraph{Limitations \& future work}
While our experiments are extensive, certain aspects remain unexplored. 
Our study focuses on identifying the single most important ViT component for adaptation rather than multiple components.
An exhaustive search would be computationally prohibitive, and a constrained search, resembling a NAS, may yield undesirable task-specific combinations \cite{zhang2022neural}.
We also did not examine how the choice of  $r$ might vary with data availability or information density; richer data may support a larger $r$ and enhance adaptability.
Lastly, \apla’s susceptibility to catastrophic forgetting remains untested --unlike adapter-based methods, which can be stored separately, \apla directly modifies the foundation model, potentially impacting retention of prior knowledge.

%% file: tables/proj_only.tex
\begin{tabular}{@{} lccccccccc @{}}
\toprule
Method & Adaptation & \birds & \cars & \aid & \isic & \vtabnatural &  \vtabspecialized & \vtabstructured & \average \\ 
\midrule
\multirow{2}{*}{\lora} & Default & 
87.5 &	93.4 &	\underline{95.4}  &	86.5 &  83.4 & 86.5	 & 63.1	& 78.2
\\
& On \wo & 
87.7  & 93.5 & 95.1 & \underline{87.9} & 83.7  & \underline{87.6}  & 62.6  & 78.3
\\
\multirow{2}{*}{\textsc{AdaptF.}} & Default &  
\textbf{88.4}	& 93.1 &	\underline{95.4} &	85.6 & 84.0 &	87.2 &	59.8 &	77.3
\\
& On \wo & 
\textbf{88.4} & 93.6 & 95.3  & 87.1  & 84.2  & 87.2  & 61.6  & 78.1 
\\
\multirow{2}{*}{\fact} & Default &  
87.8 &	93.0 &	\underline{95.4} & 85.1 & \underline{84.7} &	87.4 &	64.5 &	\underline{79.1}
\\
& On \wo & 
\underline{88.0} & \underline{93.8} & 95.3 & 86.5  & 84.4 & 87.2 & \underline{63.8} & 78.8
\\
\midrule
\textbf{\apla} & On \wo  & \underline{88.0} & \textbf{94.0} & \textbf{96.0} & \textbf{88.2} & \textbf{85.0} & \textbf{88.2} & \textbf{63.9} & \textbf{79.4} \\
\bottomrule
\end{tabular}

%% file: 7_conclusion.tex
\section{Conclusion}
\label{conclusion}

We introduced \apla, a simple yet effective method for adapting ViTs by tuning only a randomly selected subset of projection layer columns. 
Extensive experiments show that \apla achieves state-of-the-art performance while reducing computational costs, making it highly practical. 
Our results highlight that in over-parameterized models, efficiency doesn’t require added complexity -- a simple targeted re-weighting of existing features can be even more powerful.

%% file: 8_acknowledgements.tex
\paragraph{Acknowledgments.}
This work was supported by the Wallenberg AI, Autonomous Systems and Software Program (WASP). 
We acknowledge the Berzelius computational resources provided by the Knut and Alice Wallenberg Foundation at the National Supercomputer Centre and the the computational resources provided by the National Academic Infrastructure for Supercomputing in Sweden (NAISS), partially funded by the Swedish Research Council through grant agreement no. 2022-06725.

%% file: _Appendix.tex
\clearpage
\setcounter{page}{1}
\renewcommand{\thepage}{A\arabic{page}}
\maketitlesupplementary
\begin{appendices}

\noindent
We provide additional experimental details and results.
\begin{itemize}
    \item \ref{apx:more_exp_details} includes additional experimental details.
    \begin{itemize}
        \item In Section \ref{apx:more_exp_details_classification} we provide implementation details for classification tasks.
        \item In \ref{apx:more_exp_details_segmentation} we provide experimental details for semantic segmentation tasks.
        \item In \ref{apx:more_exp_details_detection} we provide experimental details for object detection and instance segmentation tasks.
    \end{itemize}
    \item Section \ref{apx:more_exp_results} includes additional experimental results.
    \begin{itemize}
        \item In \ref{apx:model_scale} we report additional results as the model's capacity increases, including both classification performance and computational cost.
        \item  In \ref{apx:more_exp_results_computes} we report additional results on the computational costs of \apla and other adaptation methods during training and inference.
        \item In \ref{apx:other_methods_wo} We examine the effect of rank $r$ on computational requirements for different low-rank methods when applied solely to \wo during training.
        \item In \ref{apx:apla_to_different_blocks} we investigate the impact of applying \apla to an increasing number of ViT blocks.
        \item In \ref{apx:rank_selection} we discuss the best $r$ values for \apla.
        \item In \ref{apx:tsne_visualizations} we visualize and discuss the quality of output features produced by different adaptation methods.
    \end{itemize}
    
\end{itemize}

\vspace{2mm}
\section{Additional Experimental Details}
\label{apx:more_exp_details}

\subsection{Image Classification}
\label{apx:more_exp_details_classification}
To ensure fair comparison against other adaptation methods, we closely follow the implementations in \cite{jia2022visual, dong2024efficient, dong2024low, jie2023fact, han20232vpt, chen2022adaptformer, hu2021lora}.
For each dataset in the general classification tasks, either the official train/val/test splits were used, or we used the splits from \cite{jia2022visual}.
For \vtab, the train/val/test splits are provided.
All models were developed in PyTorch \cite{paszke2019pytorch} and trained on Nvidia A100 GPUs using AdamW \cite{adamw} optimizer. 
Unless stated otherwise, models were trained for 100 epochs using a cosine decay learning rate schedule with a 10-epoch warm-up, following previous works \cite{jia2022visual, dong2024efficient, dong2024low, jie2023fact, han20232vpt}.
We perform grid-search to determine the hyper-parameters using the validation set of each dataset.
We also perform a grid search to determine the appropriate $r$ value in \apla for each dataset.
For \vits, we search over \( r \in \{8, 16, 128, 256, 384\} \).
For \vitb, we search over \( r \in \{8, 16, 128, 512, 768\} \).
For \vitl, we search over \( r \in \{8, 16, 128, 512, 1024\} \).
For \vitg, we search over \( r \in \{8, 16, 128, 1024, 1536\} \).

\begin{figure}[t]
\centering
\scriptsize
\begin{tabular}{@{\hspace{-2.5mm}}c@{}c@{}@{}}
\includegraphics[width=0.5\linewidth]{figures/Figure_1/fig_1_mem.pdf} &
\includegraphics[width=0.5\linewidth]{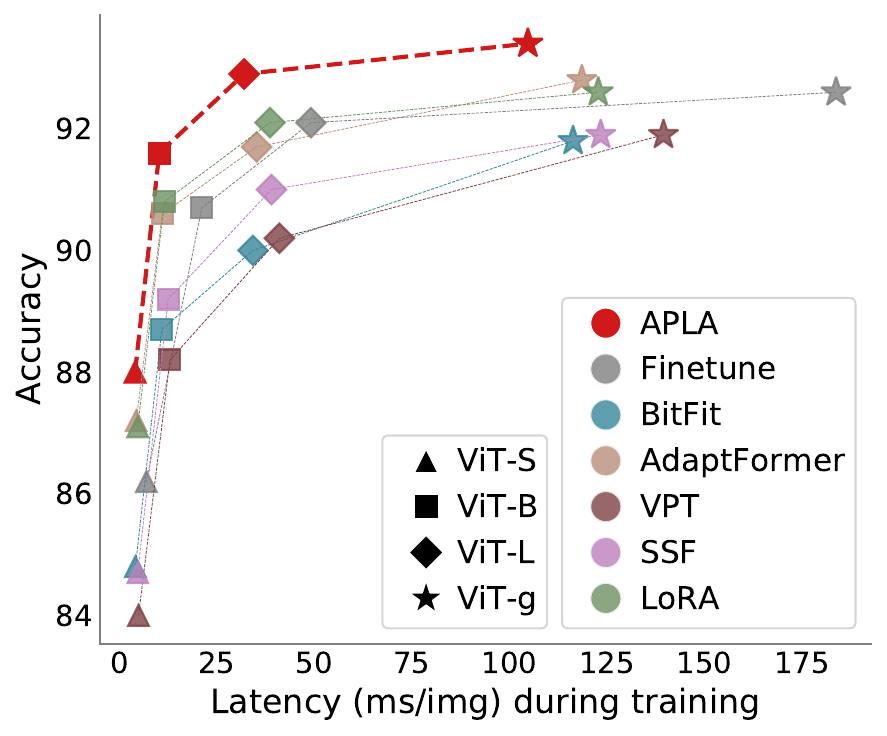} \\
\end{tabular}
\caption{
\emph{Performance vs. compute cost.} We compare each method's performance against GPU memory (left) and latency (right) during training across different model capacities. 
}
\label{fig:fig_1_mem_and_throughput}
\end{figure}

\begin{figure}[t]
    \centering
    \includegraphics[width=1.0\columnwidth]{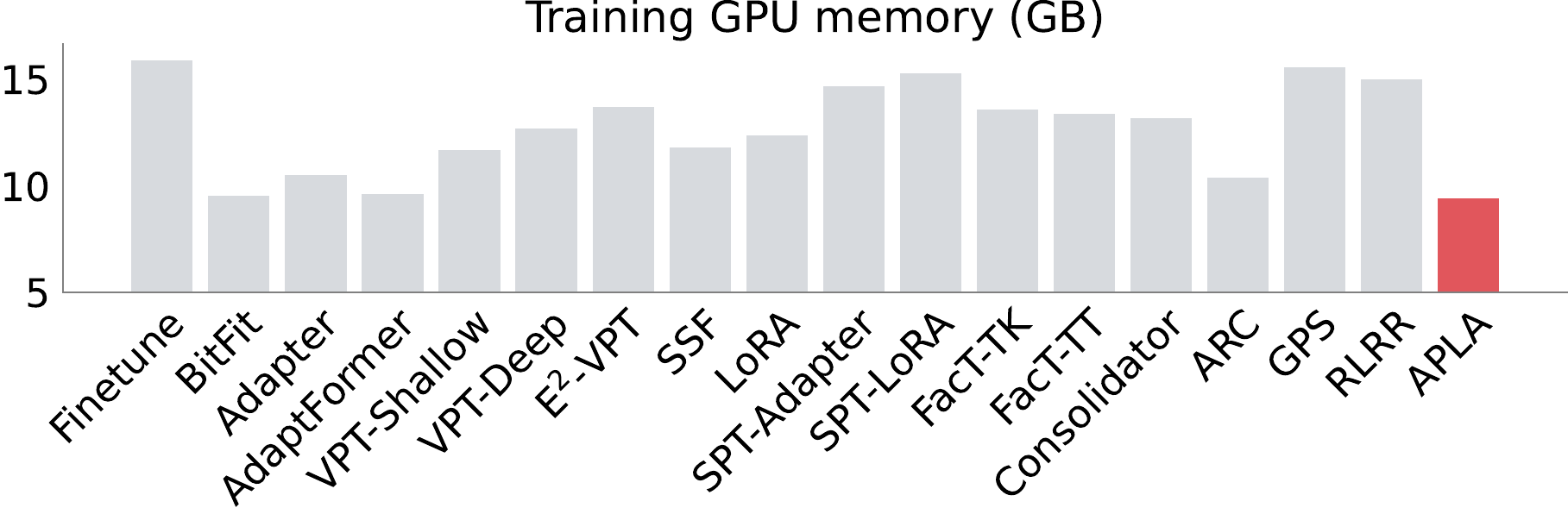}
    \includegraphics[width=1.0\columnwidth]{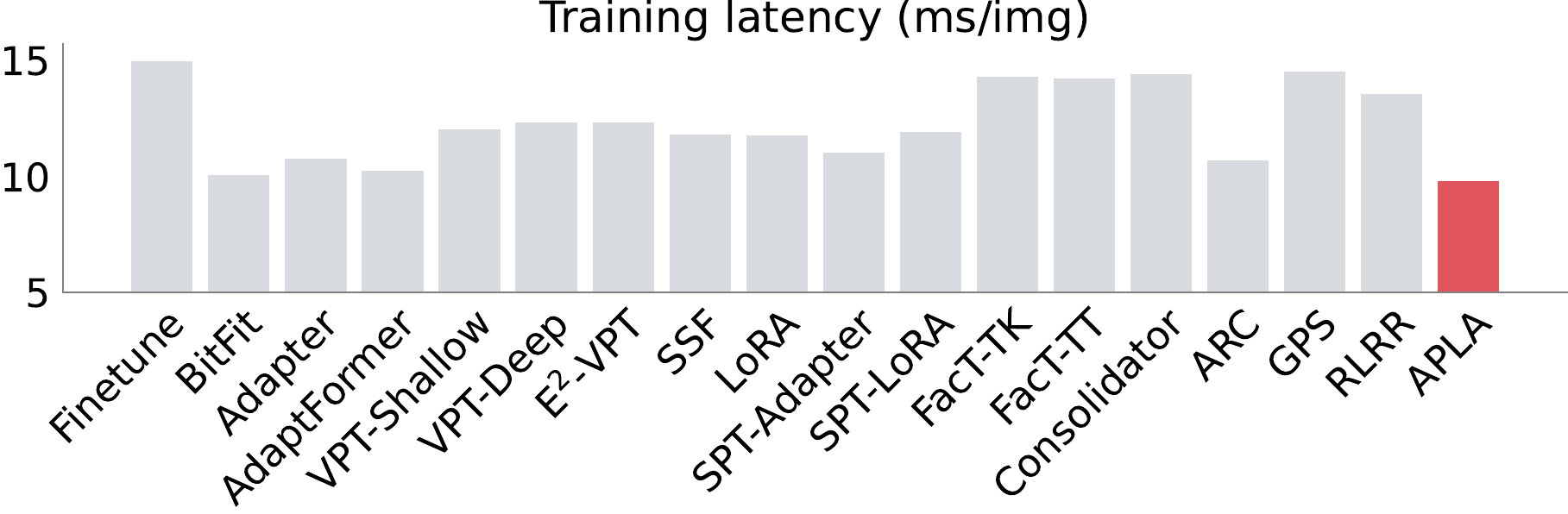}
    \\
    \vspace{2mm}
    \hrule
    \vspace{2mm}
    \includegraphics[width=1.0\columnwidth]{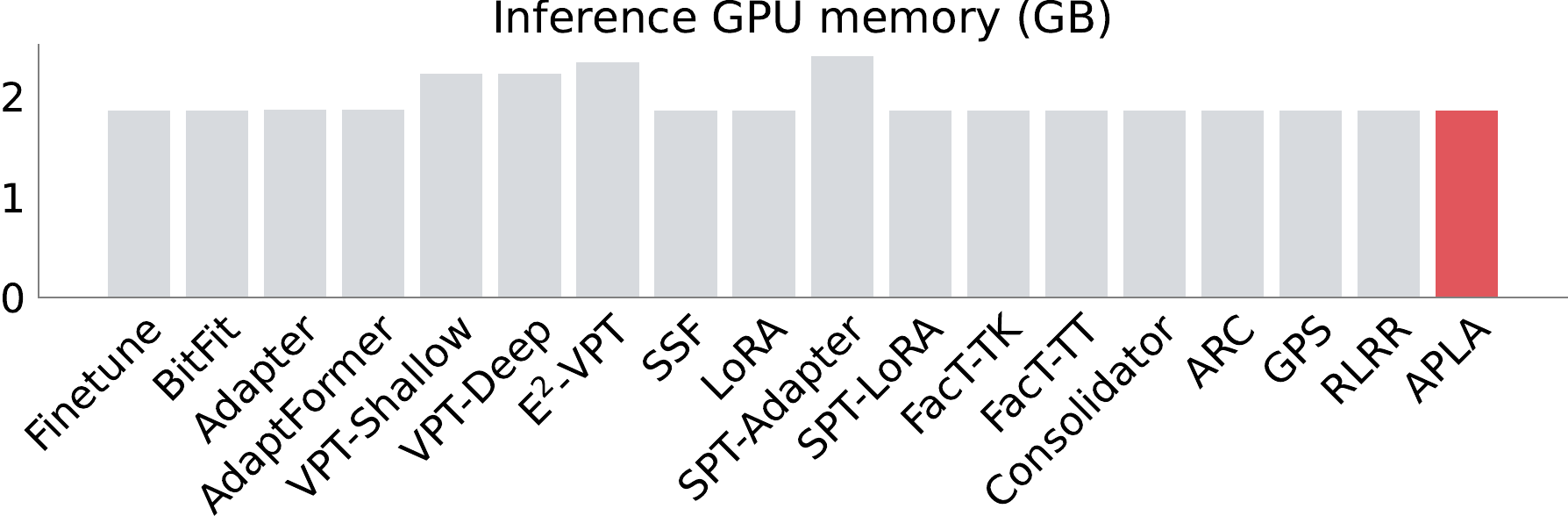}
    \includegraphics[width=1.0\columnwidth]{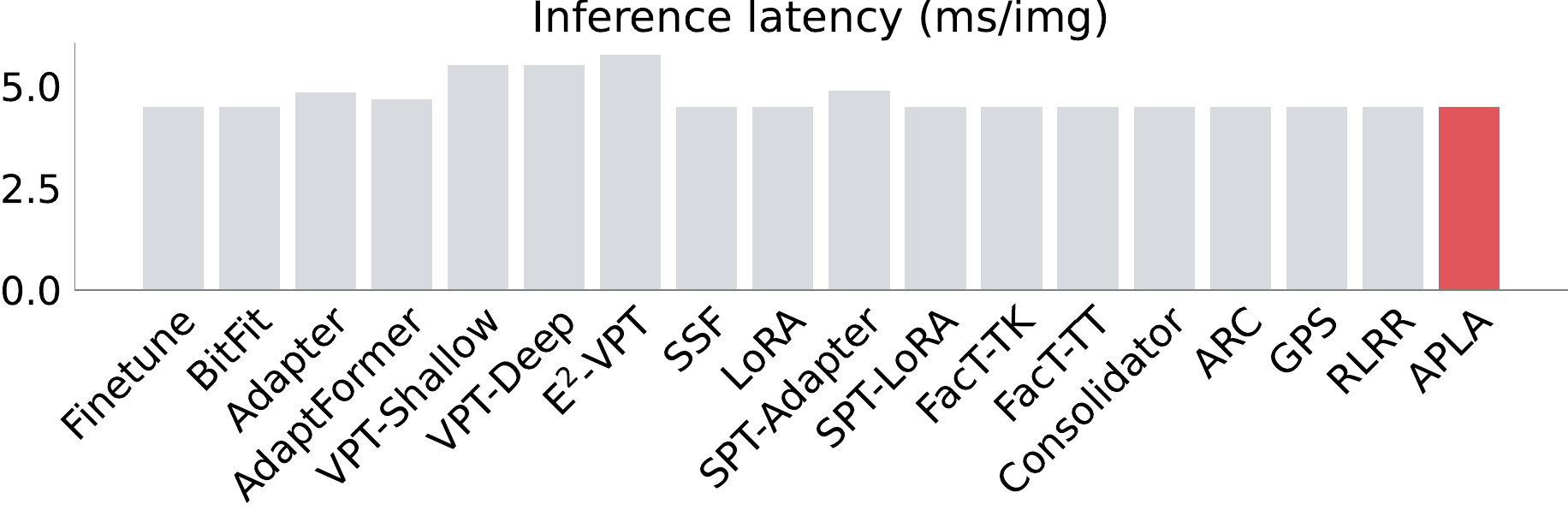}
    \\
    \vspace{2mm}
    \hrule
    \vspace{2mm}
    \includegraphics[width=1.0\columnwidth]{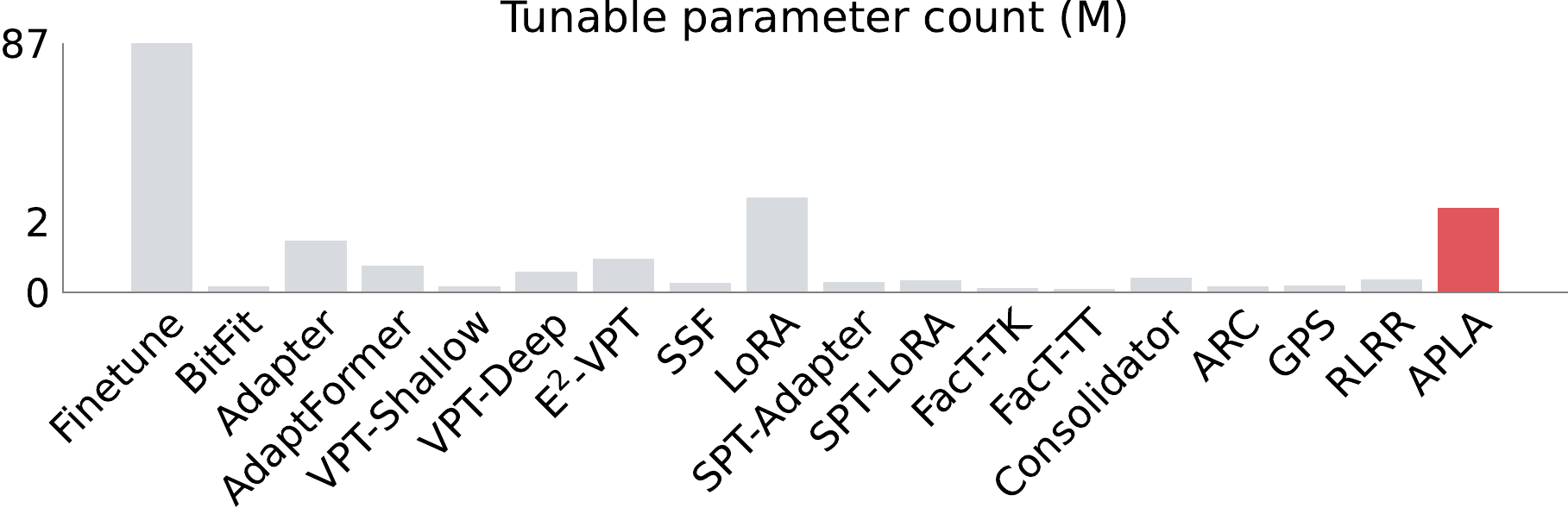}
    \caption{
    \emph{Computational costs.}
    We report memory footprint and latency of various adaptation methods during training (top) and inference (middle) for \vitb with a batch size of 64.
    Additionally, we provide the number of tunable parameter count for each method (bottom), averaged across all the datasets.
    }
    \label{fig:computes_extended}
\end{figure}

\addtolength{\tabcolsep}{-4.4pt}
\begin{table*}[t]
    \centering
    \scriptsize
    \caption{
    \emph{Classification performance across different model sizes.}  
    The \textbf{best} and \underline{second} best results are highlighted.  
    }
    \input{tables/scale}
    \vspace{2mm}
    \label{tab:scale_full}
\end{table*}
\addtolength{\tabcolsep}{+4.4pt}

\vspace{2mm}
\subsection{Semantic Segmentation}
\label{apx:more_exp_details_segmentation}

For semantic segmentation we follow \cite{jia2022visual, he2023sensitivity} and conduct experiments on the \ade dataset \cite{zhou2019semantic, zhou2017scene} using the SETR-PUP framework \cite{zheng2021rethinking} with a ViT-Large \cite{dosovitskiy2020image} model pre-trained on \imagenettwentyonek \cite{deng2009imagenet}. 
We report mean Intersection over Union (mIoU) scores for both single-scale (SS) and multi-scale (MS), following \cite{jia2022visual, he2023sensitivity}. 
Our implementation uses the \textit{mmsegmentation} \cite{mmseg2020} library.
We merely apply \apla on the default models of the library.
All training configurations are kept unchanged.

\vspace{2mm}
\subsection{Object Detection \& Instance Segmentation}
\label{apx:more_exp_details_detection}

For object detection and instance segmentation tasks we follow \cite{lian2022scaling, liu2021swin} and conduct experiments on the \coco dataset \cite{lin2014microsoft} using the Mask R-CNN framework \cite{he2017mask} with a Swin-Tiny \cite{liu2021swin} model pre-trained on \imagenetOneK \cite{deng2009imagenet}. 
We report mean Average Precision (AP) for both bounding boxes ($\mathrm{AP}^{\mathrm{bb}}$) and masks ($\mathrm{AP}^{\mathrm{m}}$) across multiple IoU thresholds and individual thresholds, following \cite{lin2014microsoft, lian2022scaling, liu2021swin}.
Our implementation uses the \textit{mmdetection} \cite{mmdetection} library.
We merely apply \apla on the default models of the library.
All training configurations are kept unchanged.

\vspace{2mm}
\section{Additional Experimental Results}
\label{apx:more_exp_results}

\subsection{Detailed Results of Different Model Scales}
\label{apx:model_scale}
To examine if \apla scales well with model size, we utilize ViT models of varying sizes (ViT-S, ViT-B, ViT-L, and ViT-g), pre-trained with \dinovtwo \cite{oquab2023dinov2}. 
Table \ref{tab:scale} in the main text shows the average results of different adaptation methods across model scales, while Table \ref{tab:scale_full} provides detailed per-dataset results. 
\apla appears to benefit from increased model capacity, performing exceptionally well with larger models.
We further present a performance-efficiency trade-off comparison in terms of GPU memory consumption and latency during training across different model sizes in Figure \ref{fig:fig_1_mem_and_throughput}.
As the model size increases, \apla outperforms all other methods both in terms of predictive performance and costs during training.

\vspace{2mm}
\subsection{Computational Costs of Adaptation Methods}
\label{apx:more_exp_results_computes}
We analyze the computational costs of adaptation methods by measuring GPU memory footprint and latency during training and inference. 
Results are shown in Figure \ref{fig:computes_extended}.
During training, \apla is the most efficient method in terms of GPU memory usage and latency and does not add any extra costs during inference.
While \apla appears to tune more parameters than other adaptation methods, one should note that parameter count alone does not necessarily reflect the true computational costs.
This inconsistency has been previously emphasized by other studies \cite{dehghani2021efficiency, cai2020tinytl}

\begin{figure}[t]
    \centering
    \includegraphics[width=1.0\columnwidth]{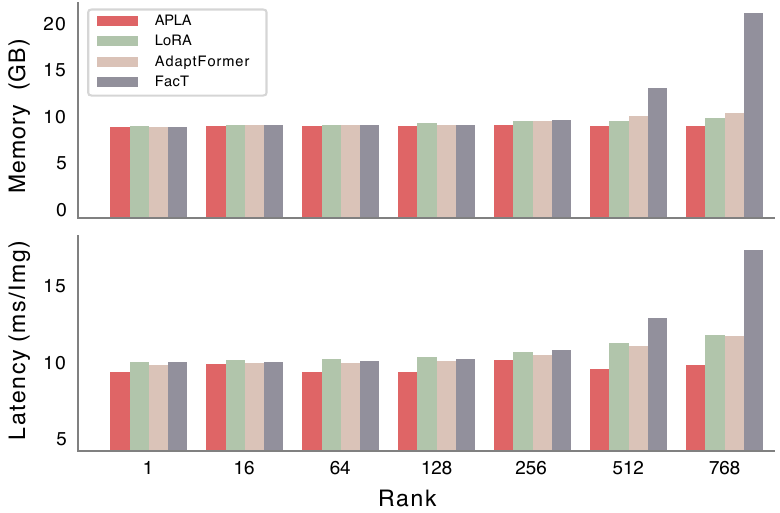}
    \caption{
    \emph{Computational requirements of different adaptation methods during training for varying $r$ values.}
    For LoRA and FacT, $r$ denotes the rank, for \apla the number of tuned columns, and for AdaptFormer, the size of the bottleneck dimension.
    }
    \label{fig:proj-only-ablation}
\end{figure}

\vspace{2mm}
\subsection{Computational Costs of Low-rank Adaptation Methods When Increasing \texorpdfstring{$r$}{r}}
\label{apx:other_methods_wo}
In Table \ref{tab:why-apla} in the main text, we investigated the impact of applying other low-rank adaptation methods on \wo, isolating the impact of the low-rank adaptation strategy.
Using the same setup, here we analyze their computational costs with respect to the choice of rank $r$, considering GPU memory and latency during training.
As shown in Figure \ref{fig:proj-only-ablation}, \apla is the only method that only minimally impacts memory and latency during, whereas other methods are affected to a larger extent as $r$ grows (\eg \fact).
Essentially, for any given rank \( r \), \apla outperforms all other low-rank adaptation methods in terms of efficiency, requiring less GPU memory and enabling faster training. 
This advantage is due to \apla's more efficient low-rank strategy and its avoidance of introducing additional trainable parameters.
This provides \apla a distinct advantage, allowing the rank to be freely adjusted for optimal results without any extra computational concerns.

\vspace{2mm}
\subsection{APLA on Increasing Number of ViT Blocks}
\label{apx:apla_to_different_blocks}

\begin{figure}
    \centering
    \includegraphics[width=0.48\columnwidth]{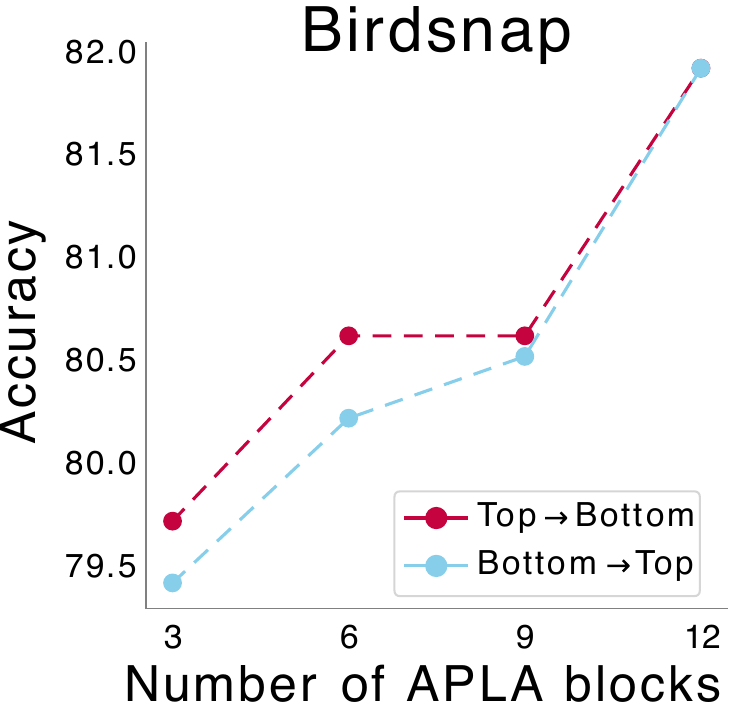}
    \hfill
    \includegraphics[width=0.48\columnwidth]{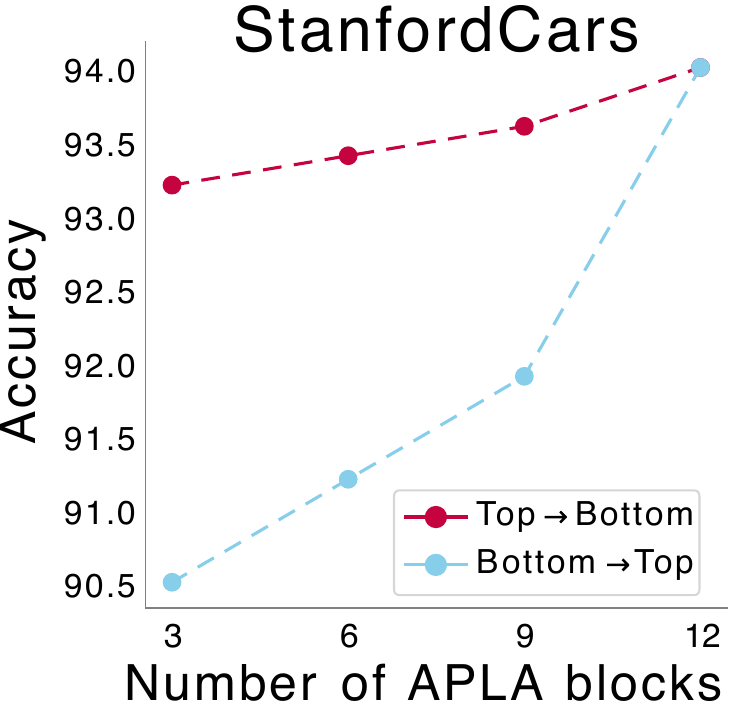}
    \\
    \caption{
    \emph{Classification performance when applying \apla to an increasing number of attention blocks.} 
    \enquote{Top-bottom} means applying \apla starting from the last ViT block and moving toward the first layer, while \enquote{"bottom-top} refers to applying it in the opposite direction.
    }
    \label{fig:selected_blocks}
\end{figure}

We investigate the impact of applying \apla to increasing number of ViT blocks, when starting from the first layer and moving towards the last layers (\enquote{Bottom $\rightarrow$ Top}) and the opposite direction (\enquote{Top $\rightarrow$ Bottom}), and present the results in Figure \ref{fig:selected_blocks}.
Applying \apla to more blocks monotonically improves performance. 
As expected, applying \apla to later transformer blocks leads to greater performance improvements than applying it to the early ViT layers.

\vspace{2mm}
\subsection{The Effect of Choice of \texorpdfstring{$r$}{r}}
\label{apx:rank_selection}
\begin{figure}[ht]
    \centering
    \includegraphics[width=1.0\linewidth]{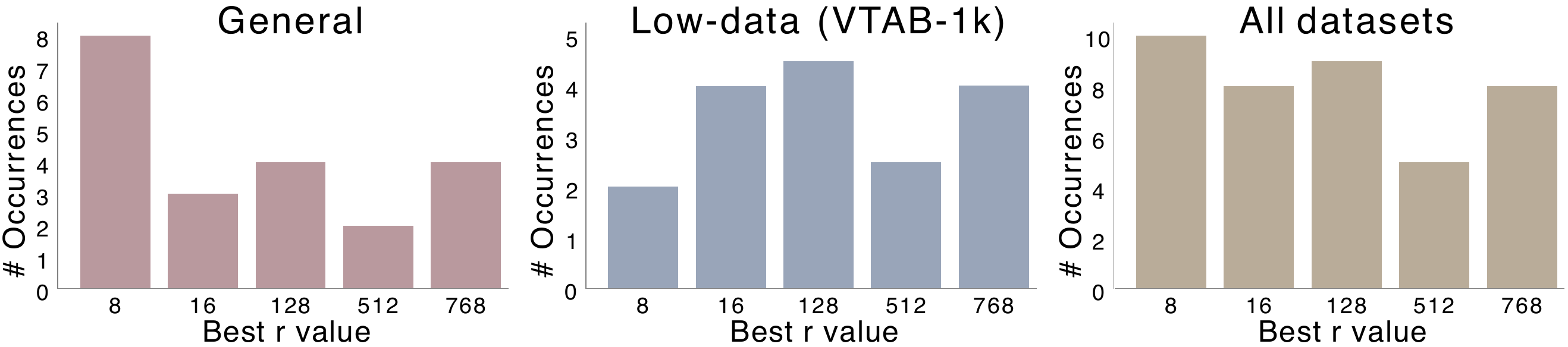}
    \caption{
    \emph{Selected $r$ values across different datasets.}
    We report the optimal $r$ values, determined by grid searches, for general classification tasks (left), datasets with limited data (middle), and all datasets (right).
    }
    \label{fig:rows-trends-summary}
\end{figure}

In APLA, the hyperparameter $r$ is used to specify how many columns of the weight matrix \wo are tuned.
In Figure \ref{fig:rows-trends-summary}, we present the selected values of $r$ for general classification tasks (left), datasets with limited data (middle), and all datasets together (right).

\vspace{2mm}
\subsection{Feature Visualization}
\label{apx:tsne_visualizations}
\begin{figure*}
\begin{minipage}{\tSNEFigureWidth}
    \input{tables/tsne/dsprites_loc_15}    
\end{minipage}
\hfill
\begin{minipage}{\tSNEFigureWidth}
    \input{tables/tsne/clevr_dist}

\end{minipage}
\\
\\
\\
\begin{minipage}{\tSNEFigureWidth}
    \input{tables/tsne/dsprites_ori}

\end{minipage}
\hfill
\begin{minipage}{\tSNEFigureWidth}
    \input{tables/tsne/svhn}

\end{minipage}
\\
\\
\\
\begin{minipage}{\tSNEFigureWidth}
    \input{tables/tsne/clevr_count}

\end{minipage}
\hfill
\begin{minipage}{\tSNEFigureWidth}
    \input{tables/tsne/eurosat}

\end{minipage}

\caption{
\emph{t-SNE visualizations.} 
We plot the t-SNE visualizations of the output \cls embeddings on \vtab using \vitb models pre-trained with \dinovtwo. 
All models have been adapted for each task. 
The numbers in parentheses indicate each adaptation method's classification performance for the task.
}
\label{fig:tsne_extra}
\end{figure*}
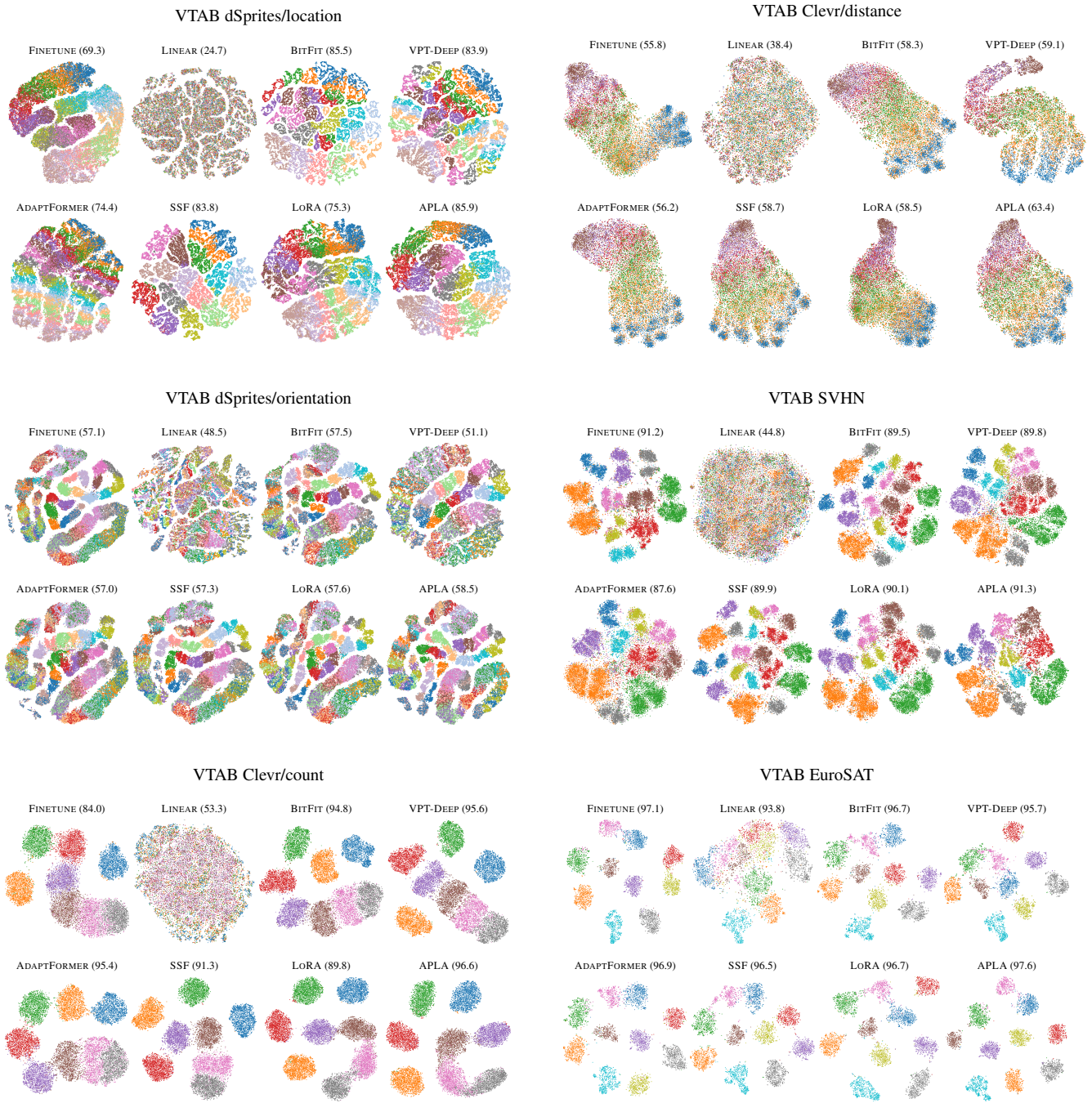

As a last sanity check, we evaluate the quality of learned representations when using \apla and compare them with those obtained from other adaptation methods, similarly to \cite{jia2022visual, lian2022scaling, chen2022adaptformer}.
In Figure \ref{fig:tsne_extra} we use t-SNE \cite{van2008visualizing} to visualize the final representations derived from the \cls token of the last ViT block for various datasets from \vtab.
Similar to other adaptation methods, \apla generates well-separated clusters for different classes, with data points from the same class positioned closely together.

\vfill\eject

\end{appendices}

%% file: tables/scale.tex
\begin{tabular}{@{} lccccccccccccccccccccccc @{}}
\toprule
{}  & 
\multicolumn{5}{c}{\textbf{\vits}} && 
\multicolumn{5}{c}{\textbf{\vitb}} && 
\multicolumn{5}{c}{\textbf{\vitl}} && 
\multicolumn{5}{c}{\textbf{\vitg}} \\
\cline{2-6}\cline{8-12}\cline{14-18}\cline{20-24}
\\
{}  & \birds & \cars & \aid & \isic &  \hspace{\avgcolspace}{\average} & \hspace{15pt} &
 \birds & \cars & \aid & \isic &  \hspace{\avgcolspace}{\average} & \hspace{15pt} &
 \birds & \cars & \aid & \isic &  \hspace{\avgcolspace}{\average} & \hspace{15pt} &
 \birds & \cars & \aid & \isic &  \hspace{\avgcolspace}{\average} \\
\midrule
\finetune & 77.5 & \textbf{91.9} & 91.7 & \underline{83.8} & \hspace{\avgcolspace} 86.2 && 85.2 & \textbf{94.4} & 95.4 & \underline{87.7} & \hspace{\avgcolspace} 90.7 && 88.2 & \underline{94.9} & 95.9 & \textbf{89.2} & \hspace{\avgcolspace} \underline{92.1} && 90.0 & \underline{95.2} & 96.5 & 88.8 & \hspace{\avgcolspace} 92.6
\\
\midrule
\linear & 81.3 & 83.1 & 88.5 & 51.5 & \hspace{\avgcolspace} 76.1 && 86.6 & 88.4 & 91.2 & 55.3 & \hspace{\avgcolspace} 80.4 && 89.1 & 89.8 & 93.4 & 62.1 & \hspace{\avgcolspace} 83.6 && 90.2 & 91.0 & 93.6 & 67.3 & \hspace{\avgcolspace} 85.5 
\\
\mlp & 80.6 & 83.1 & 88.8 & 71.1 & \hspace{\avgcolspace} 80.9 && 86.4 & 88.3 & 91.6 & 71.9 & \hspace{\avgcolspace} 84.6 && 88.9 & 89.9 & 92.9 & 73.0 & \hspace{\avgcolspace} 86.2 && 89.9 & 91.0 & 93.1 & 77.1 & \hspace{\avgcolspace} 87.8 
\\
\partialadapt & 81.1 & 83.1 & 88.3 & 50.6 & \hspace{\avgcolspace} 75.8 && 86.5 & 88.1 & 90.9 & 56.1 & \hspace{\avgcolspace} 80.4 && 89.1 & 89.8 & 93.3 & 62.0 & \hspace{\avgcolspace} 83.6 && 90.3 & 91.0 & 93.6 & 65.7 & \hspace{\avgcolspace} 85.2 
\\
\midrule
\bias & 83.1 & 89.7 & 93.1 & 73.2 & \hspace{\avgcolspace} 84.8 &&  87.9 & 92.5 & 95.2 & 79.0 & \hspace{\avgcolspace} 88.7 && 90.4 & 93.8 & 95.8 & 80.1 & \hspace{\avgcolspace} 90.0 && 90.8 & 94.5 & 95.9 & 85.8 & \hspace{\avgcolspace} 91.8 
\\
\adapter & \underline{83.2} & 91.3 & \underline{93.3} & \underline{83.8} & \hspace{\avgcolspace} \underline{87.9} &&  \textbf{88.4} & 93.5 & 95.0 & 84.3 & \hspace{\avgcolspace} 90.3 && 90.4 & 94.8 & 95.8 & 86.4 & \hspace{\avgcolspace} 91.9 && 91.1 & 95.1 & 96.3 & 86.8 & \hspace{\avgcolspace} 92.3 
\\
\adaptformer & \textbf{83.6} & 90.6 & 92.8 & 81.9 & \hspace{\avgcolspace} 87.2 && \textbf{88.4} & 93.1 & 95.4 & 85.6 & \hspace{\avgcolspace} 90.6 && \textbf{90.8} & 94.2 & 95.8 & 86.0 & \hspace{\avgcolspace} 91.7 && 91.5 & 94.9 & 96.0 & \underline{88.9} & \hspace{\avgcolspace} \underline{92.8} 
\\
\vptshallow & 81.7 & 86.3 & 91.3 & 72.0 & \hspace{\avgcolspace} 82.8 && 86.7 & 90.6 & 91.6 & 76.5 & \hspace{\avgcolspace} 86.4 && 89.0 & 91.6 & 93.0 & 79.7 & \hspace{\avgcolspace} 88.3 && 89.8 & 92.1 & 95.1 & 78.7 & \hspace{\avgcolspace} 88.9 
\\
\vptdeep & 80.1 & 86.8 & 92.5 & 76.7 & \hspace{\avgcolspace} 84.0 && 87.3 & 91.5 & 94.4 & 79.6 & \hspace{\avgcolspace} 88.2 && 89.1 & 93.4 & 95.7 & 82.6 & \hspace{\avgcolspace} 90.2 && 91.1 & 94.5 & 96.2 & 85.9 & \hspace{\avgcolspace} 91.9 
\\
\etwovpt & 80.0 & 87.6 & 91.9 & 77.3 & \hspace{\avgcolspace} 84.2 && 86.6 & 91.2 & 93.7 & 80.9 & \hspace{\avgcolspace} 88.1 && 89.5 & 93.8 & 95.8 & 84.3 & \hspace{\avgcolspace} 90.9 && 91.2 & 94.5 & 95.9 & 85.3 & \hspace{\avgcolspace} 91.7 
\\
\ssf & 81.7 & 89.5 & 92.7 & 74.9 & \hspace{\avgcolspace} 84.7 && 88.1 & 92.7 & 95.3 & 80.7 & \hspace{\avgcolspace} 89.2 && \underline{90.6} & 94.0 & 95.9 & 83.5 & \hspace{\avgcolspace} 91.0 && 91.0 & 94.5 & 95.9 & 86.3 & \hspace{\avgcolspace} 91.9 
\\
\lora & 80.8 & 91.0 & \underline{93.3} & 83.2 & \hspace{\avgcolspace} 87.1 && 87.9 & 93.4 & 95.4 & 86.5 & \hspace{\avgcolspace} \underline{90.8} && 89.9 & 94.8 & 95.9 & \underline{87.9} & \hspace{\avgcolspace} \underline{92.1} && 89.8 & 94.8 & \underline{96.7} & \underline{88.9} & \hspace{\avgcolspace} 92.6 
\\
\sptadapter & 83.0 & 91.0 & \underline{93.3} & 81.4 & \hspace{\avgcolspace} 87.2 && 88.1 & 93.1 & \underline{95.6} & 82.1 & \hspace{\avgcolspace} 89.7 && \underline{90.6} & 93.4 & 95.6 & 80.5 & \hspace{\avgcolspace} 90.0 && 90.7 & 93.2 & 95.0 & 79.0 & \hspace{\avgcolspace} 89.5 
\\
\sptlora & 82.8 & 91.1 & 93.2 & 83.2 & \hspace{\avgcolspace} 87.6 && 87.9 & 92.8 & 95.4 & 82.2 & \hspace{\avgcolspace} 89.6 && 89.6 & 93.8 & 95.5 & 82.9 & \hspace{\avgcolspace} 90.5 && 90.3 & 93.5 & 95.2 & 78.5 & \hspace{\avgcolspace} 89.4 
\\
\facttk & 82.5 & 90.6 & 92.9 & 81.7 & \hspace{\avgcolspace} 86.9 && 87.8 & 93.0 & 95.4 & 85.1 & \hspace{\avgcolspace} 90.3 && \underline{90.6} & 94.5 & \underline{96.3} & 84.4 & \hspace{\avgcolspace} 91.5 && 91.6 & 95.1 & 96.1 & 85.3 & \hspace{\avgcolspace} 92.0 
\\
\facttt & 82.3 & 89.7 & 91.6 & 78.0 & \hspace{\avgcolspace} 85.4 && 87.6 & 92.9 & 94.5 & 81.5 & \hspace{\avgcolspace} 89.1 && 90.5 & 94.6 & \underline{96.3} & 83.7 & \hspace{\avgcolspace} 91.3 && 91.4 & 94.7 & 96.0 & 84.1 & \hspace{\avgcolspace} 91.6 
\\
\arc & 82.9 & 89.4 & 93.2 & 78.5 & \hspace{\avgcolspace} 86.0 && \underline{88.2} & 92.6 & 95.6 & 82.5 & \hspace{\avgcolspace} 89.7 && \underline{90.6} & 94.3 & 95.5 & 83.9 & \hspace{\avgcolspace} 91.1 && 91.5 & 94.7 & 95.7 & 85.5 & \hspace{\avgcolspace} 91.9 
\\
\rlrr & 82.4 & 89.5 & \underline{93.3} & 72.2 & \hspace{\avgcolspace} 84.4 && 87.9 & 92.4 & 95.0 & 81.7 & \hspace{\avgcolspace} 89.3 && \underline{90.6} & 94.0 & 95.7 & 86.7 & \hspace{\avgcolspace} 91.8 && \underline{91.6} & 94.6 & 95.7 & 87.2 & \hspace{\avgcolspace} 92.3 
\\
\midrule
\apla & 82.4 & \underline{91.4} & \textbf{93.5} & \textbf{84.5} & \hspace{\avgcolspace} \textbf{88.0} && 88.0 & \underline{94.0} & \textbf{96.0} & \textbf{88.2} & \hspace{\avgcolspace} \textbf{91.6} && \underline{90.6} & \textbf{95.1} & \textbf{96.5} & \textbf{89.2} & \hspace{\avgcolspace} \textbf{92.9} && \textbf{91.7} & \textbf{95.4} & \textbf{96.8} & \textbf{89.5} & \hspace{\avgcolspace} \textbf{93.4} 
\\
\bottomrule

\end{tabular}

%% file: tables/tsne/dsprites_loc_15.tex
\setlength{\tabcolsep}{\tSNEColSpace}
\begin{tabular}{@{} ccccc @{}}
    \multicolumn{4}{c}{\footnotesize{VTAB dSprites/location}} \\
    \addlinespace[0.1cm]
    \tSNETitle{\finetune (69.3)} &
    \tSNETitle{\linear  (24.7)} & 
    \tSNETitle{\bias  (85.5)} & 
    \tSNETitle{\vptdeep  (83.9)} 
    \\
    \includegraphics[width=\tSNEPanelWidth]{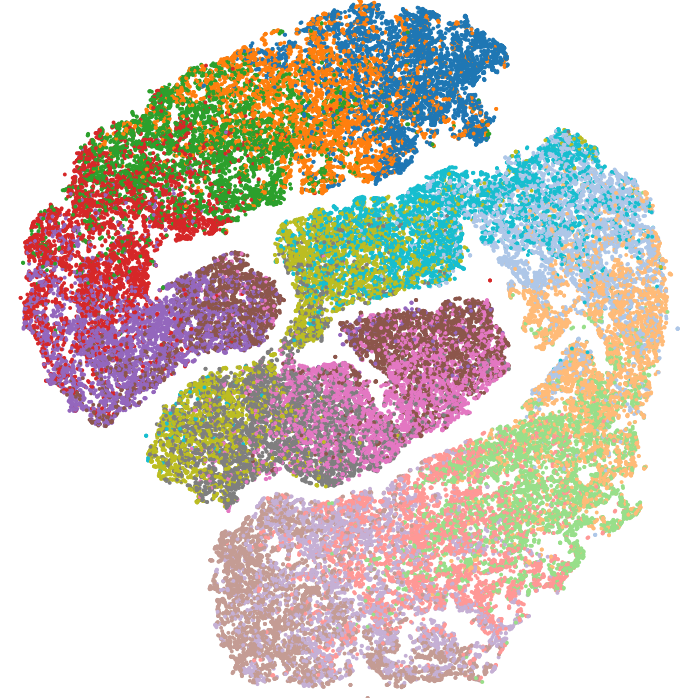} & 
    \includegraphics[width=\tSNEPanelWidth]{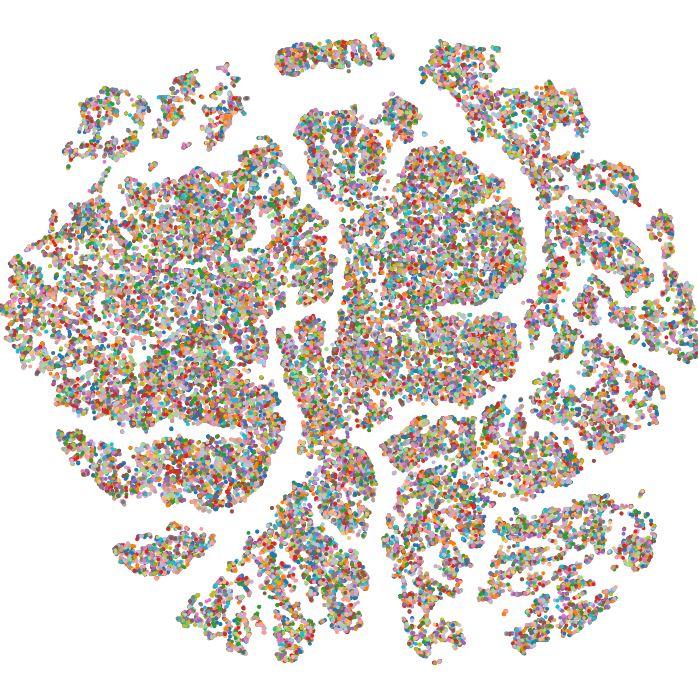} & 
    \includegraphics[width=\tSNEPanelWidth]{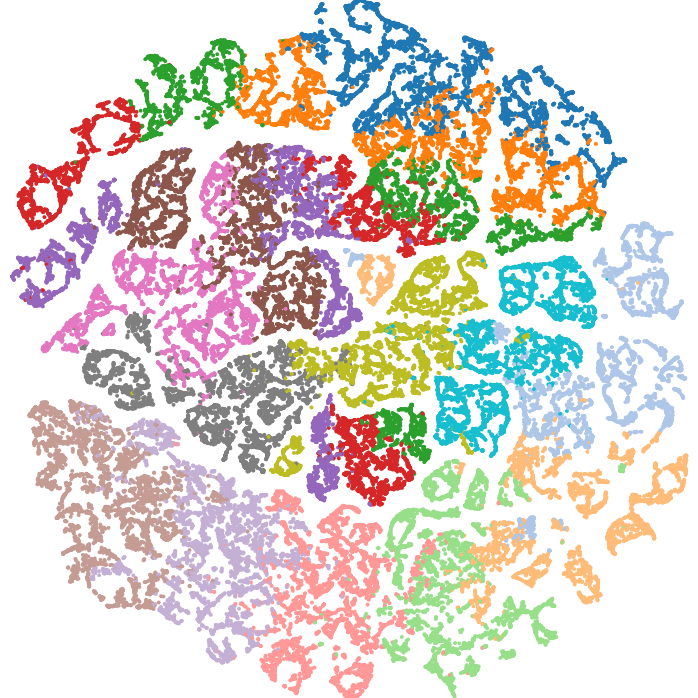} & 
    \includegraphics[width=\tSNEPanelWidth]{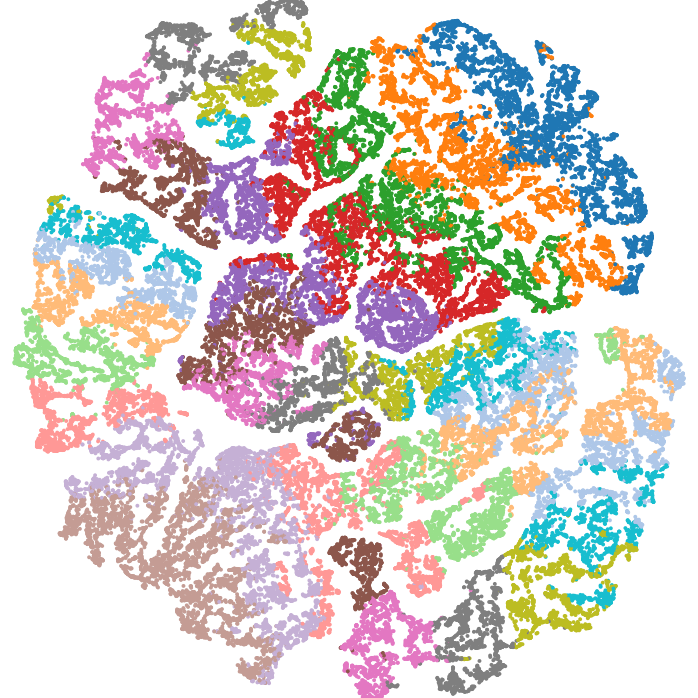} 
    \\
    \tSNETitle{\adaptformer  (74.4)} & 
    \tSNETitle{\ssf  (83.8)} & 
    \tSNETitle{\lora (75.3)} & 
    \tSNETitle{\apla  (85.9)} 
    \\
    \includegraphics[width=\tSNEPanelWidth]{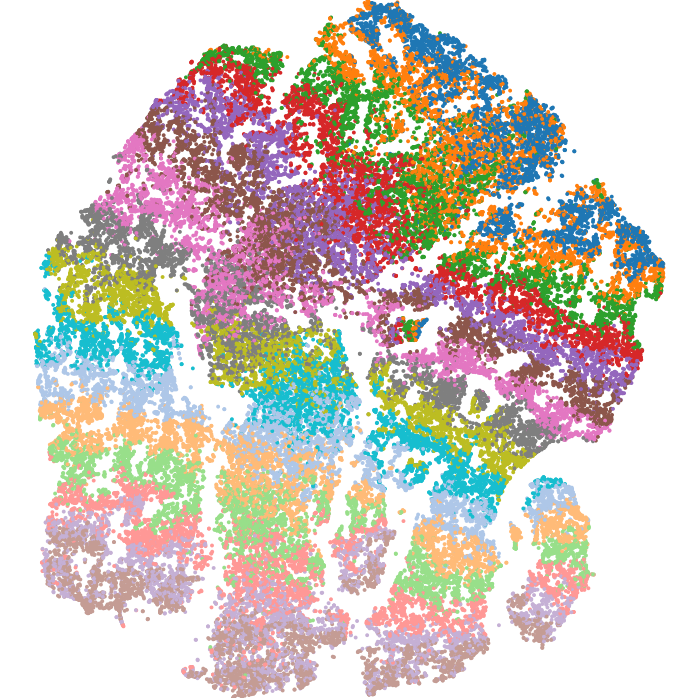} & 
    \includegraphics[width=\tSNEPanelWidth]{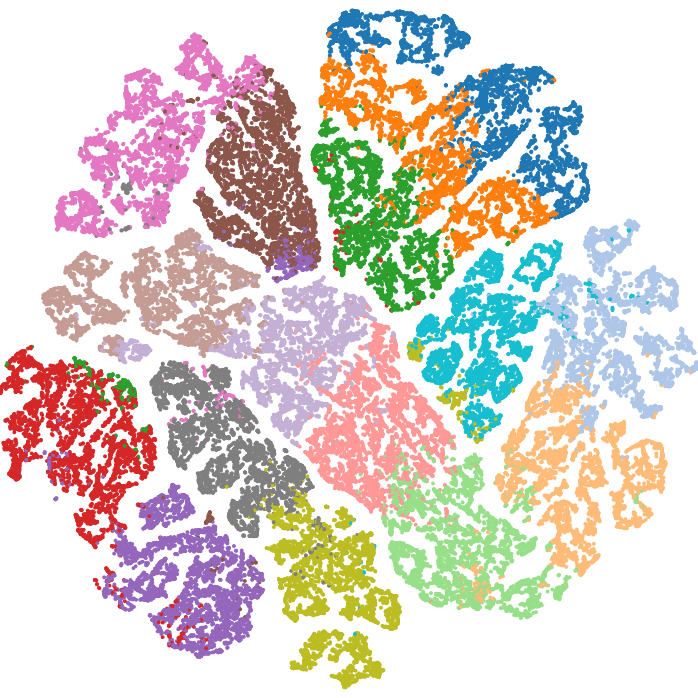} & 
    \includegraphics[width=\tSNEPanelWidth]{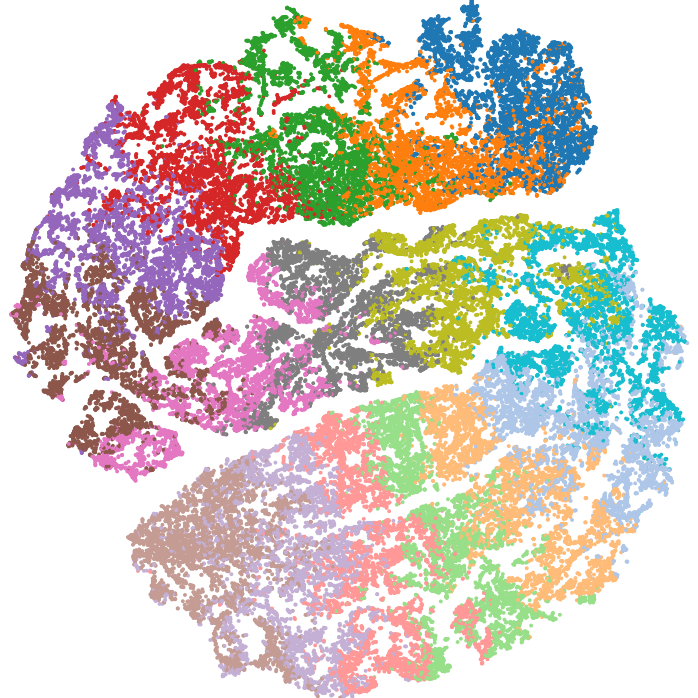} & 
    \includegraphics[width=\tSNEPanelWidth]{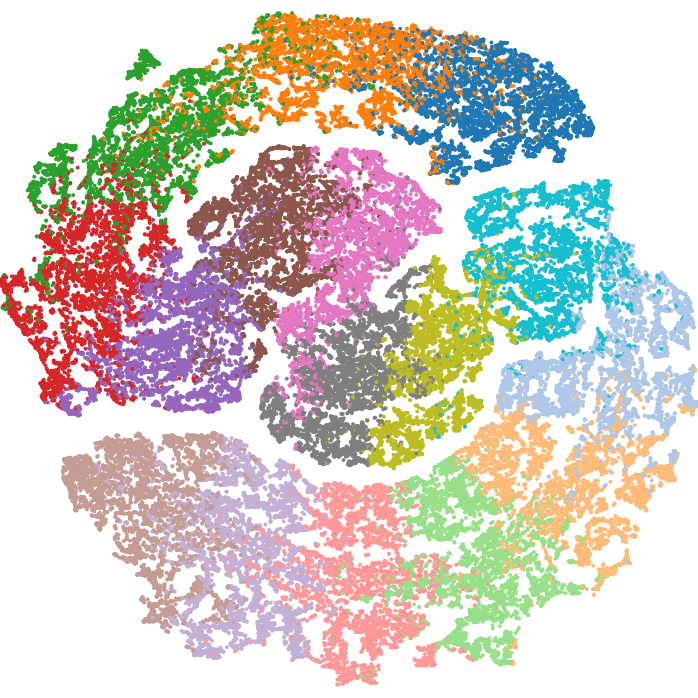} \\
\end{tabular}

%% file: tables/tsne/clevr_dist.tex
\setlength{\tabcolsep}{\tSNEColSpace}
\begin{tabular}{@{} ccccc @{}}
    \multicolumn{4}{c}{\footnotesize{VTAB Clevr/distance}} \\
    \addlinespace[0.1cm]
    \tSNETitle{\finetune (55.8)} &
    \tSNETitle{\linear  (38.4)} & 
    \tSNETitle{\bias  (58.3)} & 
    \tSNETitle{\vptdeep  (59.1)} 
    \\
    \includegraphics[width=0.25\columnwidth]{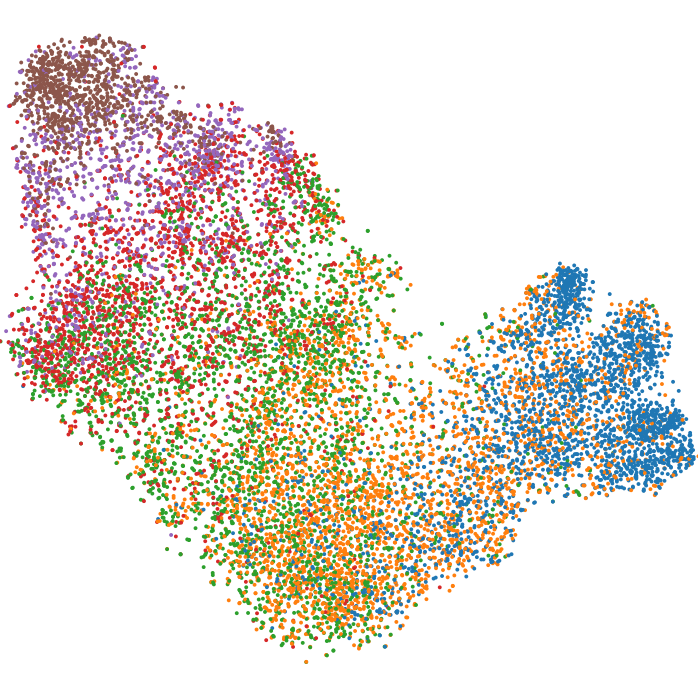} & 
    \includegraphics[width=0.25\columnwidth]{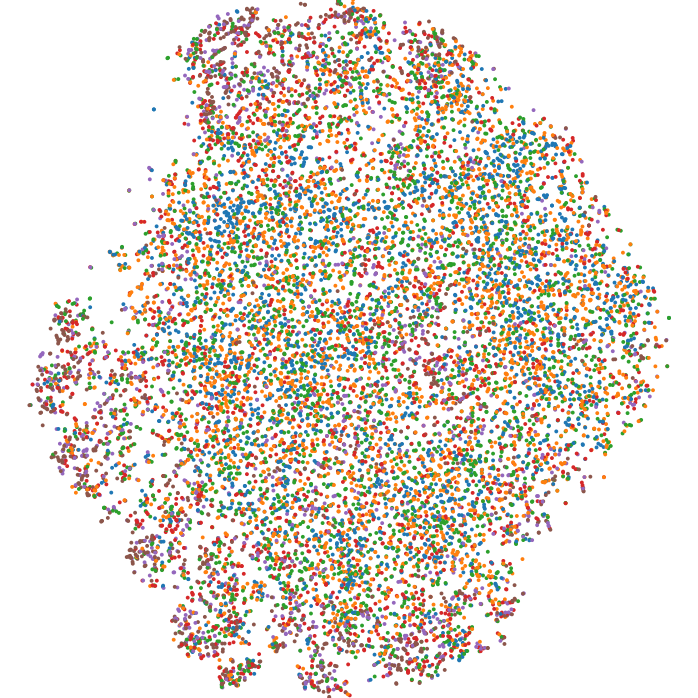} & 
    \includegraphics[width=0.25\columnwidth]{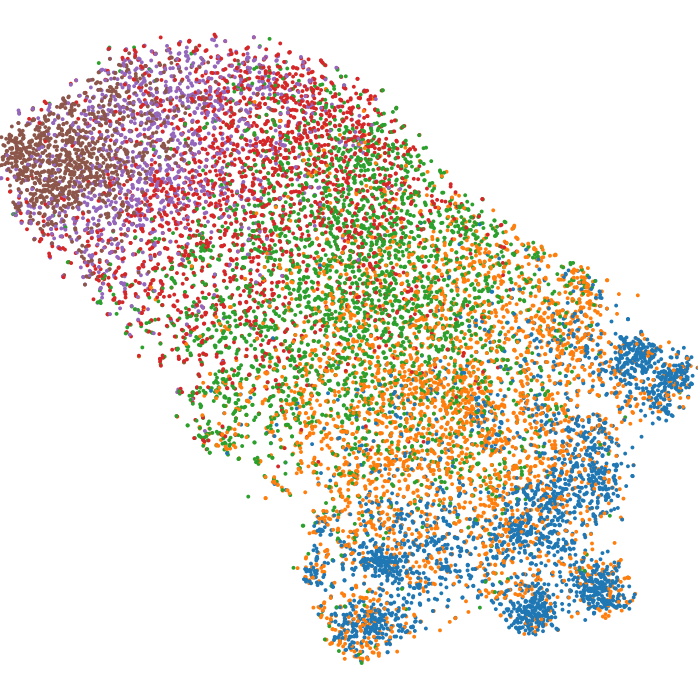} & 
    \includegraphics[width=0.25\columnwidth]{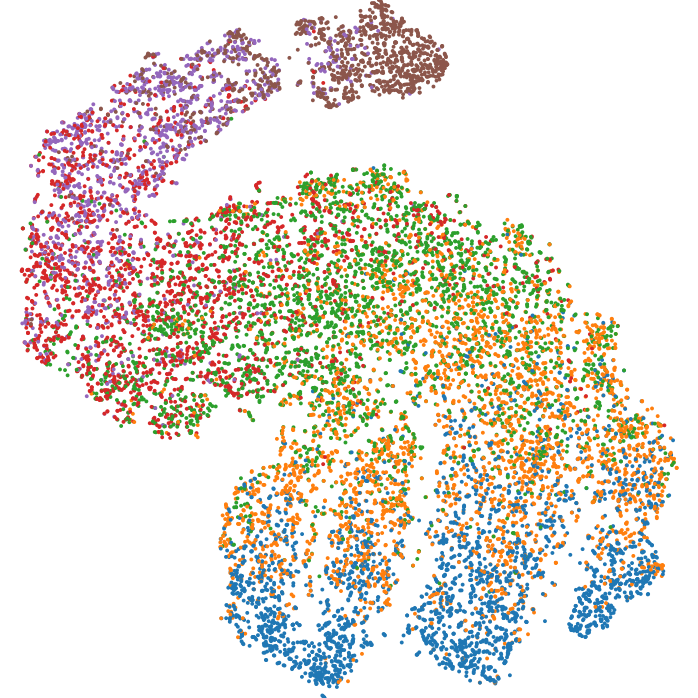} 
    \\
    \tSNETitle{\adaptformer  (56.2)} & 
    \tSNETitle{\ssf  (58.7)} & 
    \tSNETitle{\lora (58.5)} & 
    \tSNETitle{\apla  (63.4)} 
    \\
    \includegraphics[width=0.25\columnwidth]{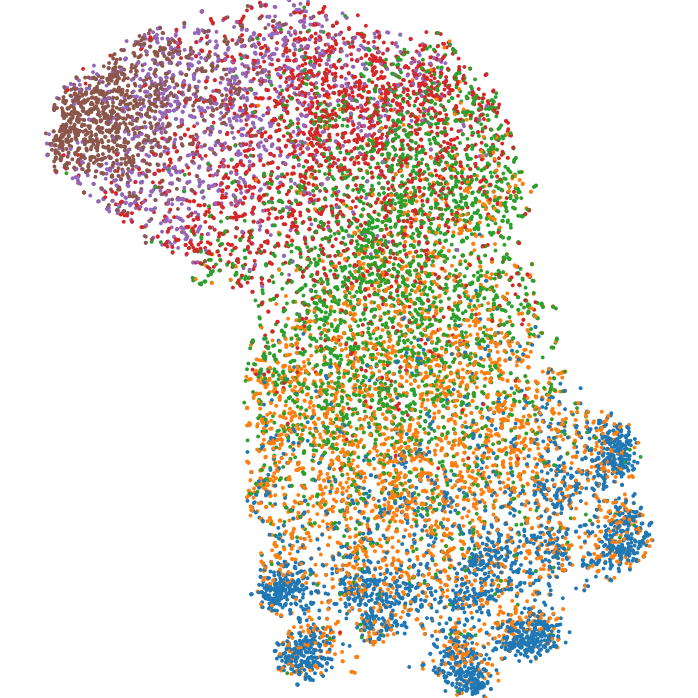} & 
    \includegraphics[width=0.25\columnwidth]{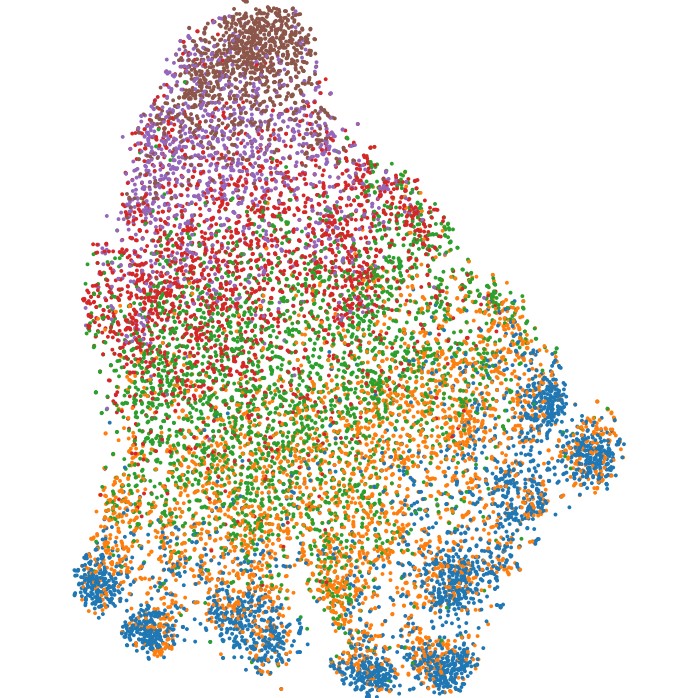} & 
    \includegraphics[width=0.25\columnwidth]{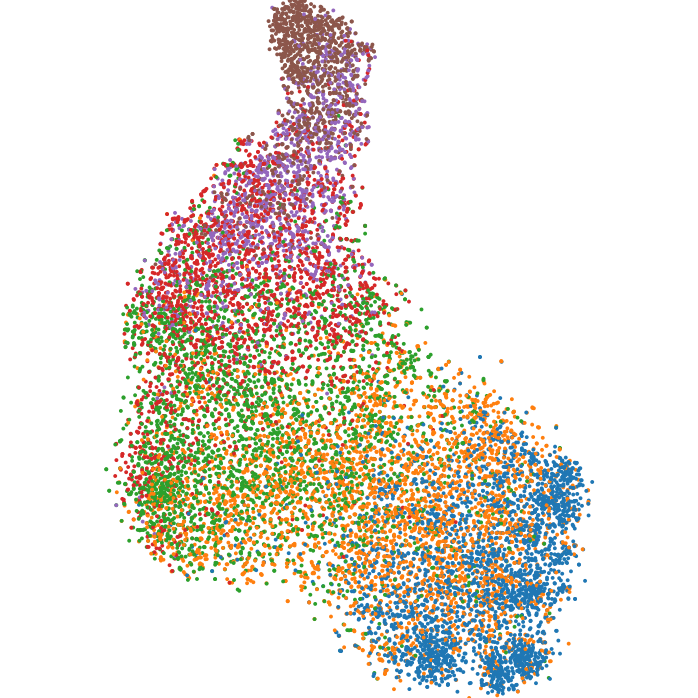} & 
    \includegraphics[width=0.25\columnwidth]{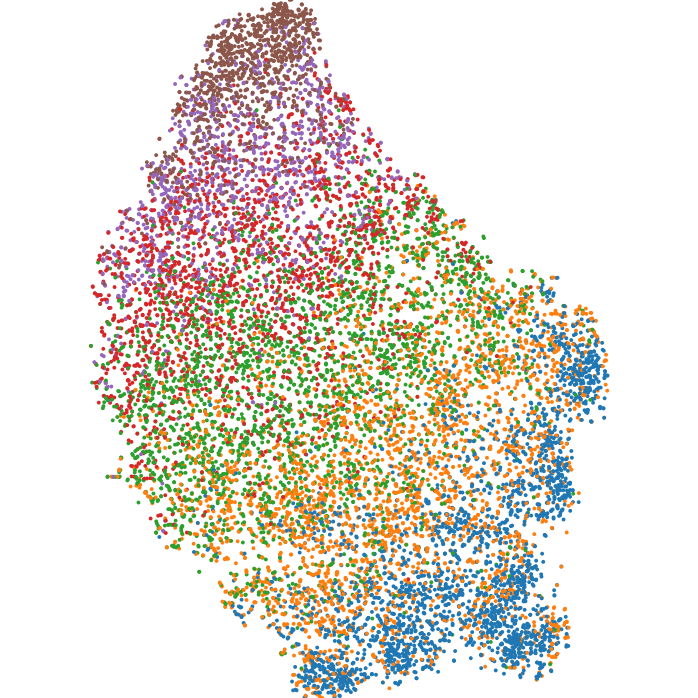} \\
\end{tabular}

%% file: tables/tsne/dsprites_ori.tex
\setlength{\tabcolsep}{\tSNEColSpace}
\begin{tabular}{@{} ccccc @{}}
    \multicolumn{4}{c}{\footnotesize{VTAB dSprites/orientation}} \\
    \addlinespace[0.1cm]
    \tSNETitle{\finetune (57.1)} &
    \tSNETitle{\linear  (48.5)} & 
    \tSNETitle{\bias  (57.5)} & 
    \tSNETitle{\vptdeep  (51.1)} 
    \\
    \includegraphics[width=\tSNEPanelWidth]{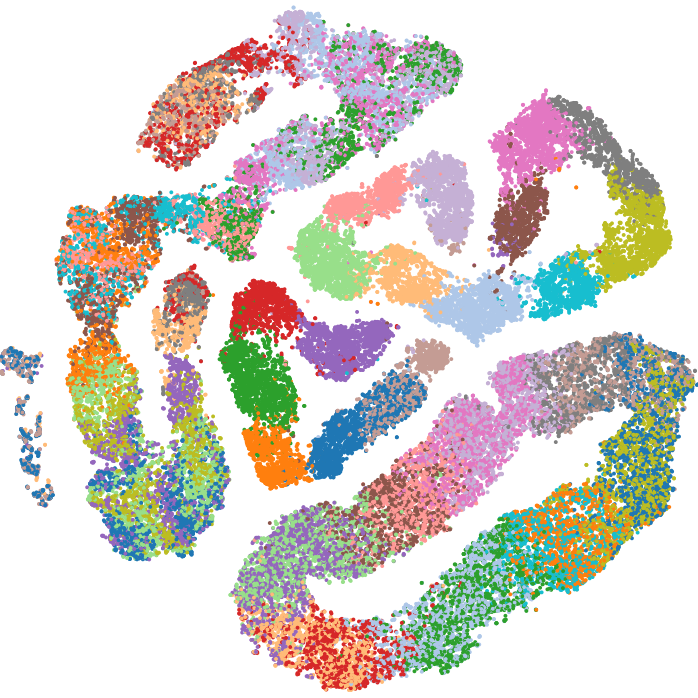} & 
    \includegraphics[width=\tSNEPanelWidth]{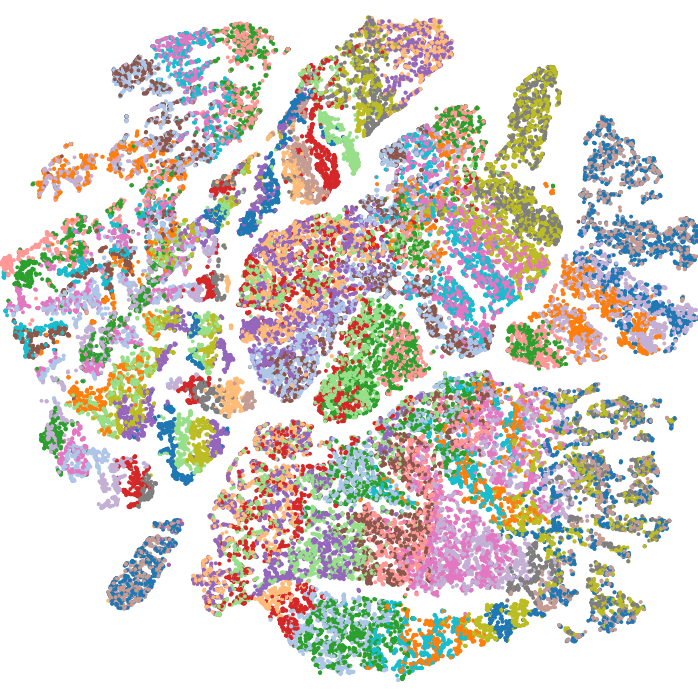} & 
    \includegraphics[width=\tSNEPanelWidth]{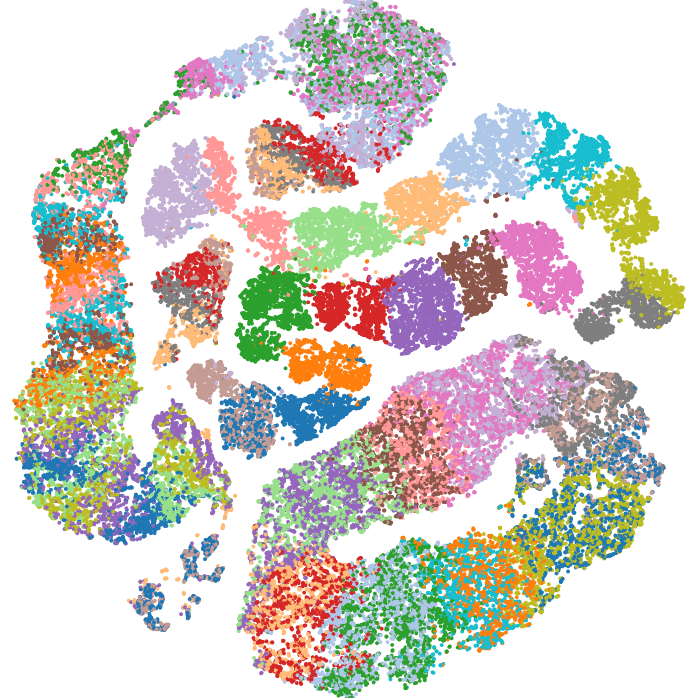} & 
    \includegraphics[width=\tSNEPanelWidth]{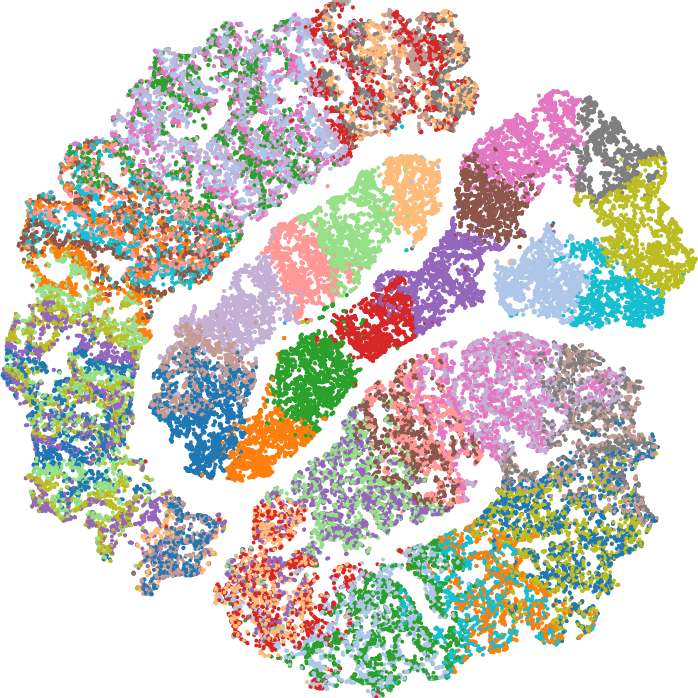} 
    \\
    \tSNETitle{\adaptformer  (57.0)} & 
    \tSNETitle{\ssf  (57.3)} & 
    \tSNETitle{\lora (57.6)} & 
    \tSNETitle{\apla  (58.5)} 
    \\
    \includegraphics[width=\tSNEPanelWidth]{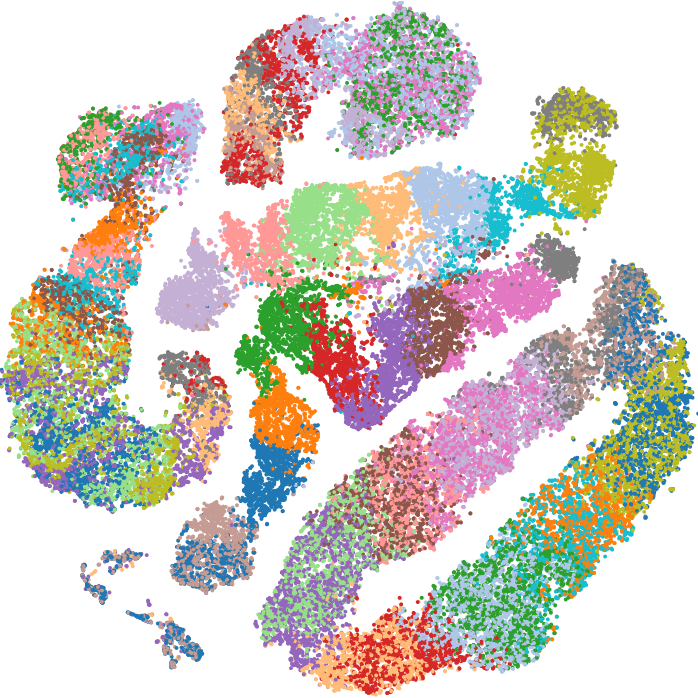} & 
    \includegraphics[width=\tSNEPanelWidth]{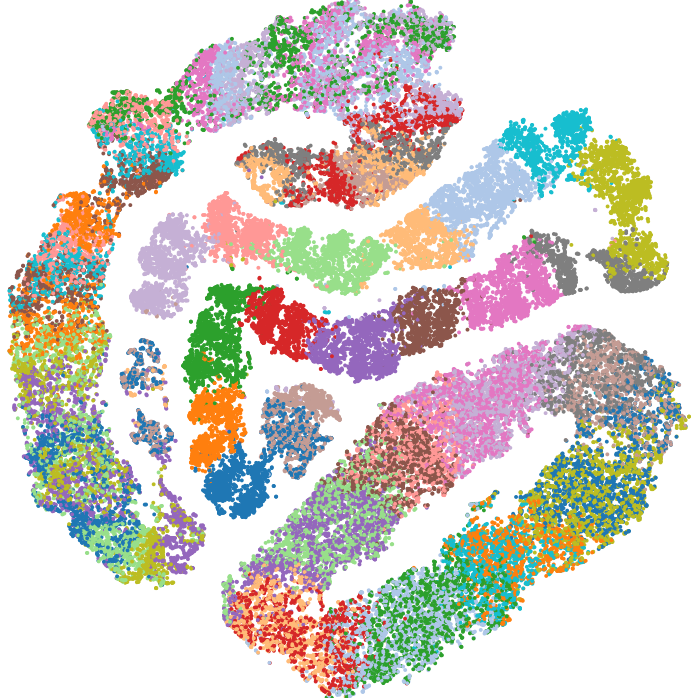} & 
    \includegraphics[width=\tSNEPanelWidth]{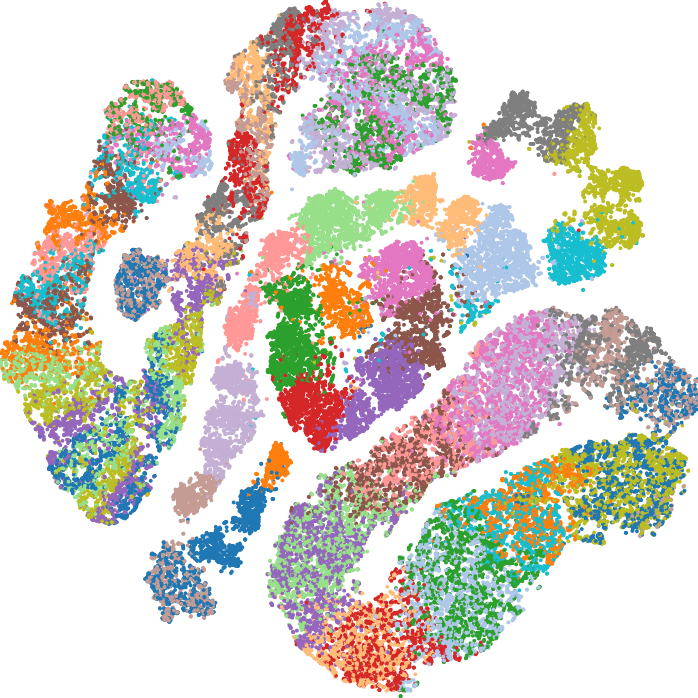} & 
    \includegraphics[width=\tSNEPanelWidth]{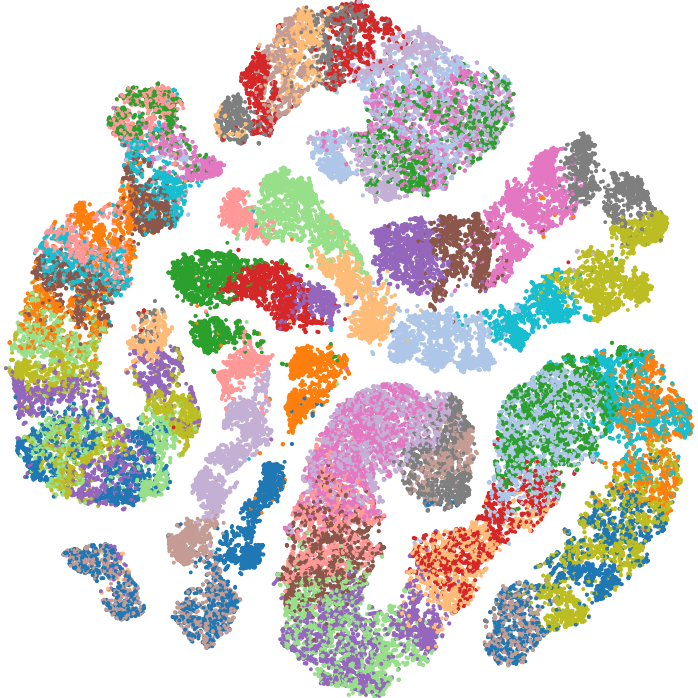} \\
\end{tabular}

%% file: tables/tsne/svhn.tex
\setlength{\tabcolsep}{\tSNEColSpace}
\begin{tabular}{@{} ccccc @{}}
    \multicolumn{4}{c}{\footnotesize{VTAB SVHN}} \\
    \addlinespace[0.1cm]
    \tSNETitle{\finetune (91.2)} &
    \tSNETitle{\linear  (44.8)} & 
    \tSNETitle{\bias  (89.5)} & 
    \tSNETitle{\vptdeep  (89.8)} 
    \\
    \includegraphics[width=\tSNEPanelWidth]{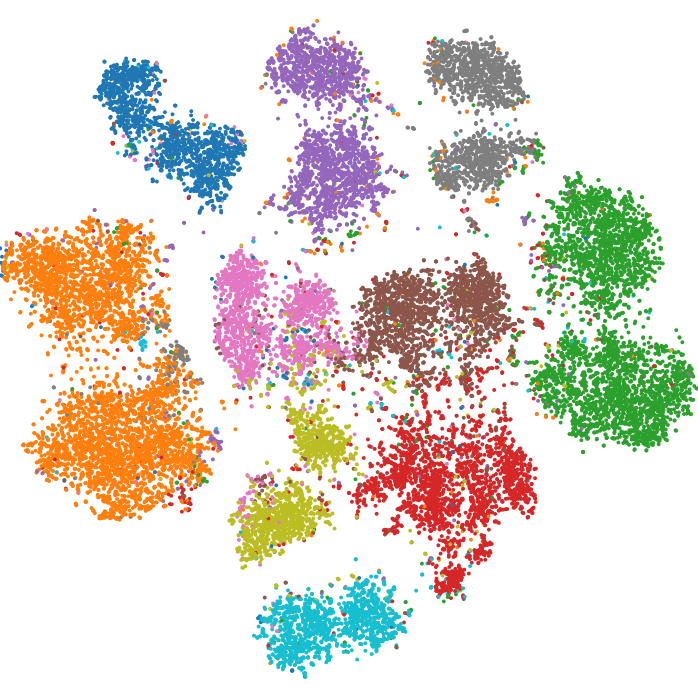} & 
    \includegraphics[width=\tSNEPanelWidth]{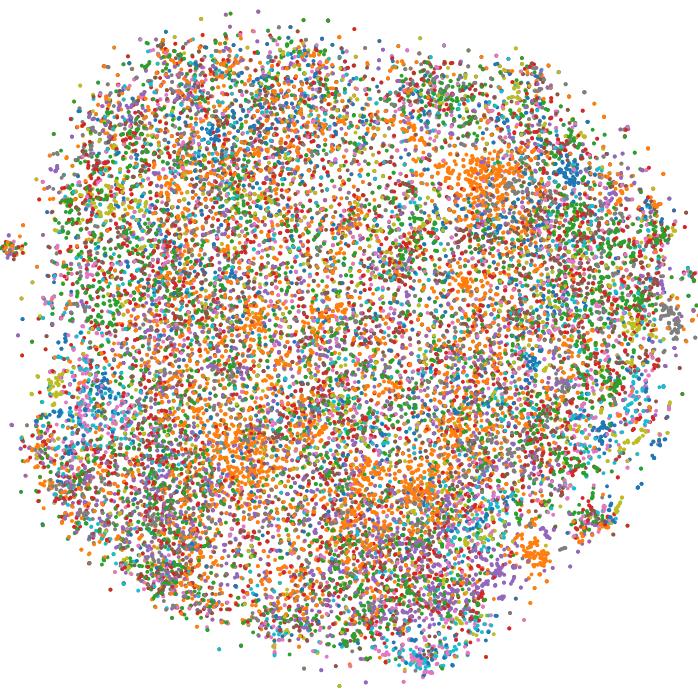} & 
    \includegraphics[width=\tSNEPanelWidth]{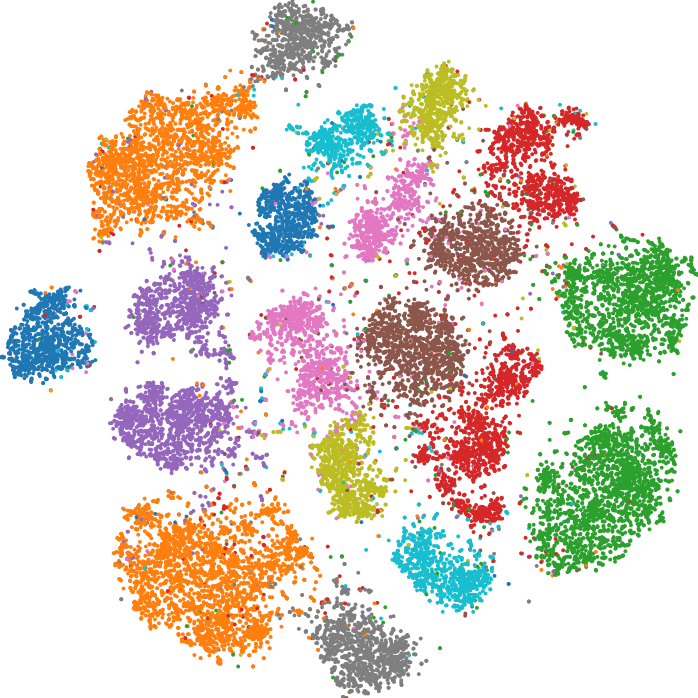} & 
    \includegraphics[width=\tSNEPanelWidth]{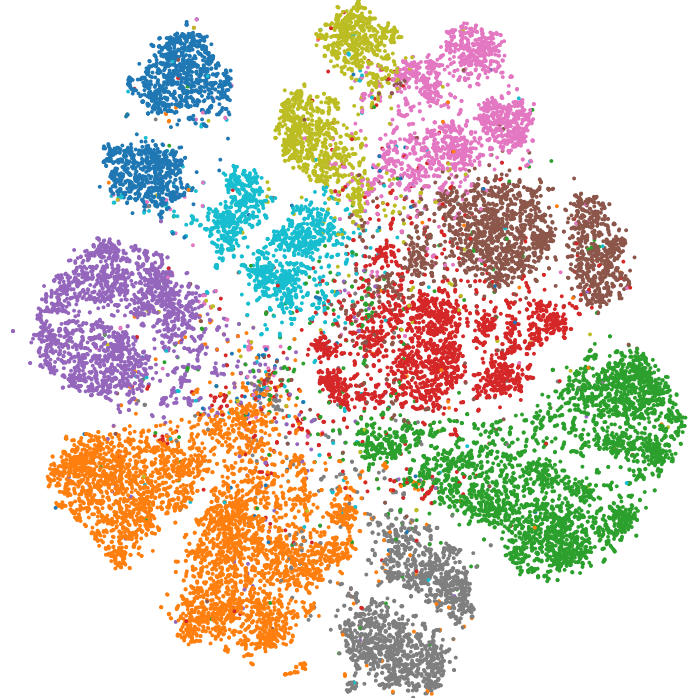} 
    \\
    \tSNETitle{\adaptformer  (87.6)} & 
    \tSNETitle{\ssf  (89.9)} & 
    \tSNETitle{\lora (90.1)} & 
    \tSNETitle{\apla  (91.3)} 
    \\
    \includegraphics[width=\tSNEPanelWidth]{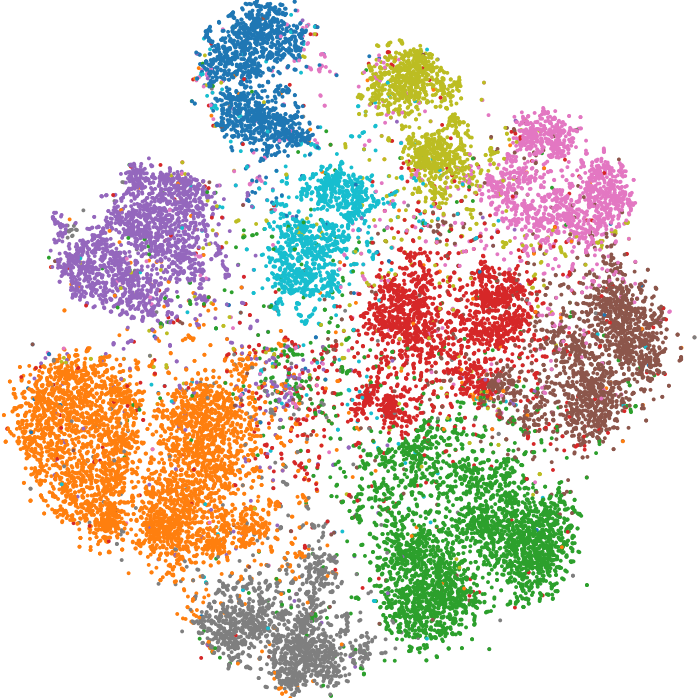} & 
    \includegraphics[width=\tSNEPanelWidth]{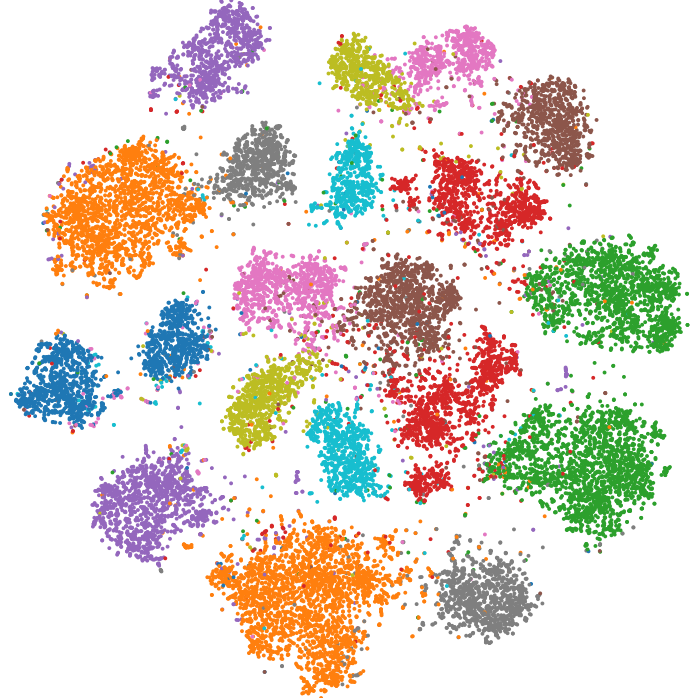} & 
    \includegraphics[width=\tSNEPanelWidth]{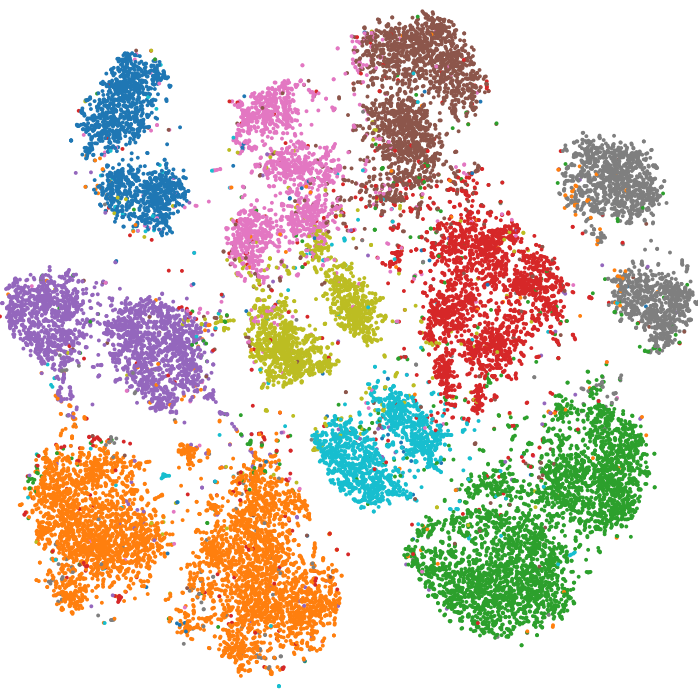} & 
    \includegraphics[width=\tSNEPanelWidth]{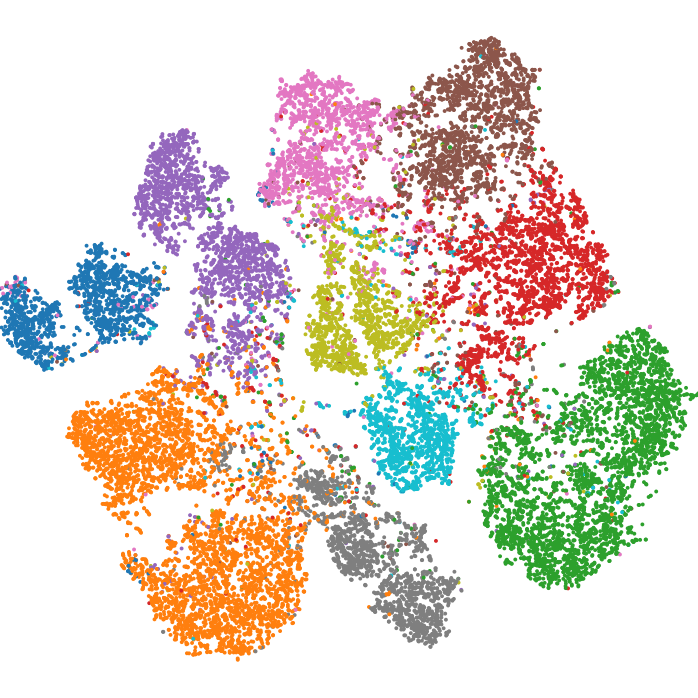} \\
\end{tabular}

%% file: tables/tsne/clevr_count.tex
\setlength{\tabcolsep}{\tSNEColSpace}
\begin{tabular}{@{} ccccc @{}}
    \multicolumn{4}{c}{\footnotesize{VTAB Clevr/count}} \\
    \addlinespace[0.1cm]
    \tSNETitle{\finetune (84.0)} &
    \tSNETitle{\linear  (53.3)} & 
    \tSNETitle{\bias  (94.8)} & 
    \tSNETitle{\vptdeep  (95.6)} 
    \\
    \includegraphics[width=\tSNEPanelWidth]{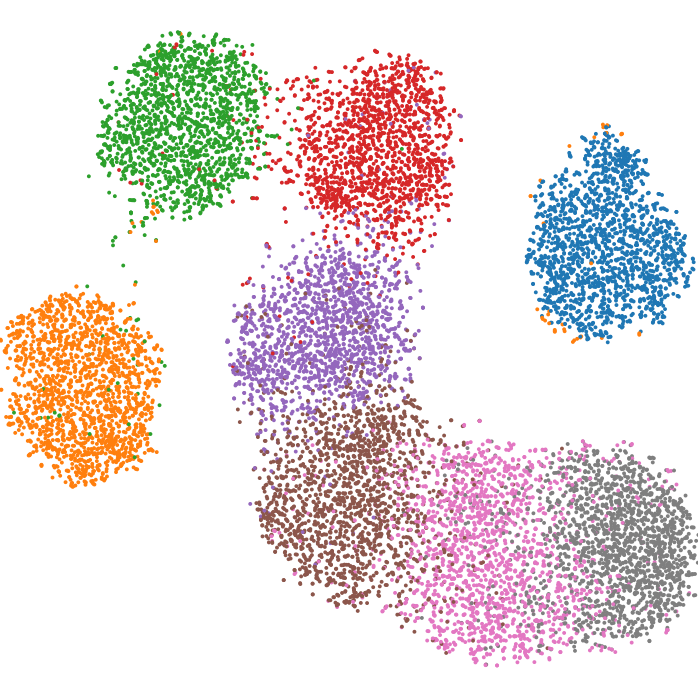} & 
    \includegraphics[width=\tSNEPanelWidth]{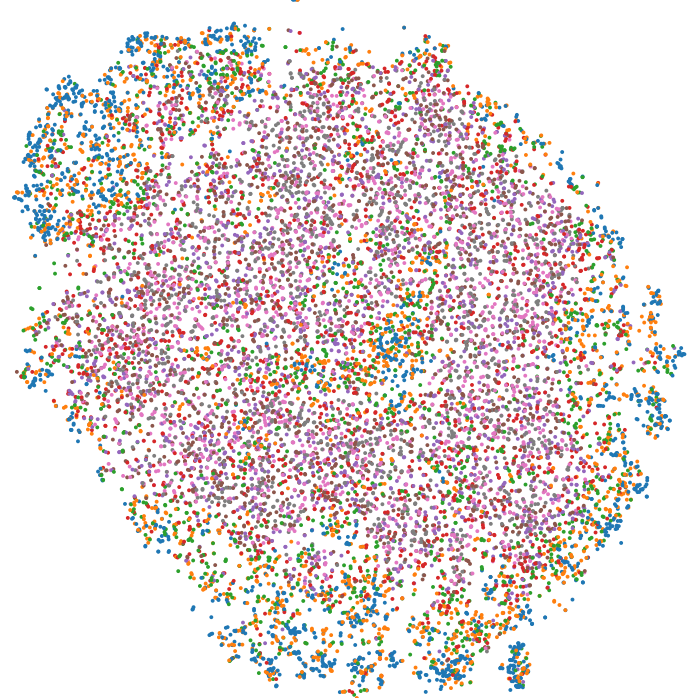} & 
    \includegraphics[width=\tSNEPanelWidth]{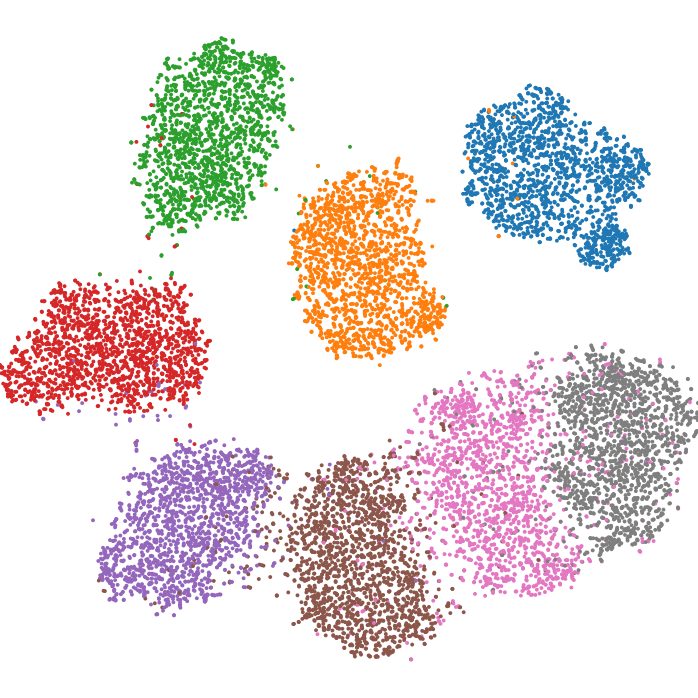} & 
    \includegraphics[width=\tSNEPanelWidth]{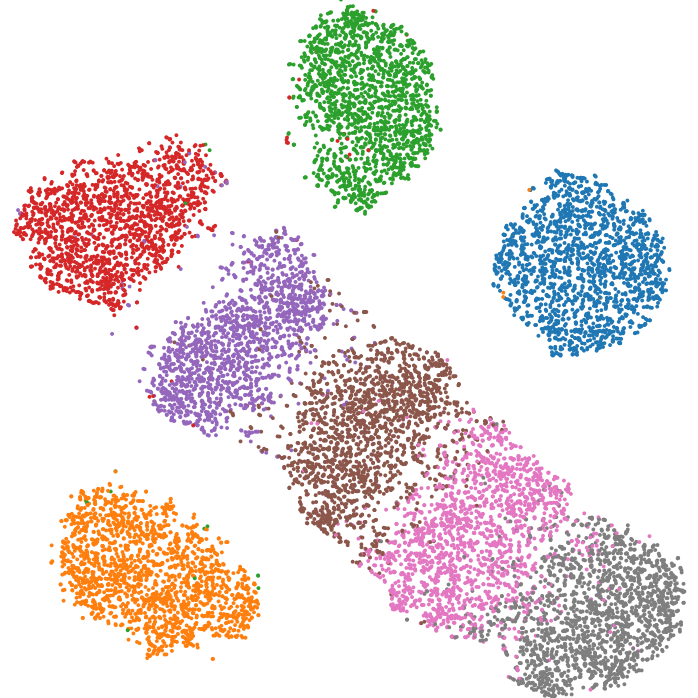} 
    \\
    \tSNETitle{\adaptformer  (95.4)} & 
    \tSNETitle{\ssf  (91.3)} & 
    \tSNETitle{\lora (89.8)} & 
    \tSNETitle{\apla  (96.6)} 
    \\
    \includegraphics[width=\tSNEPanelWidth]{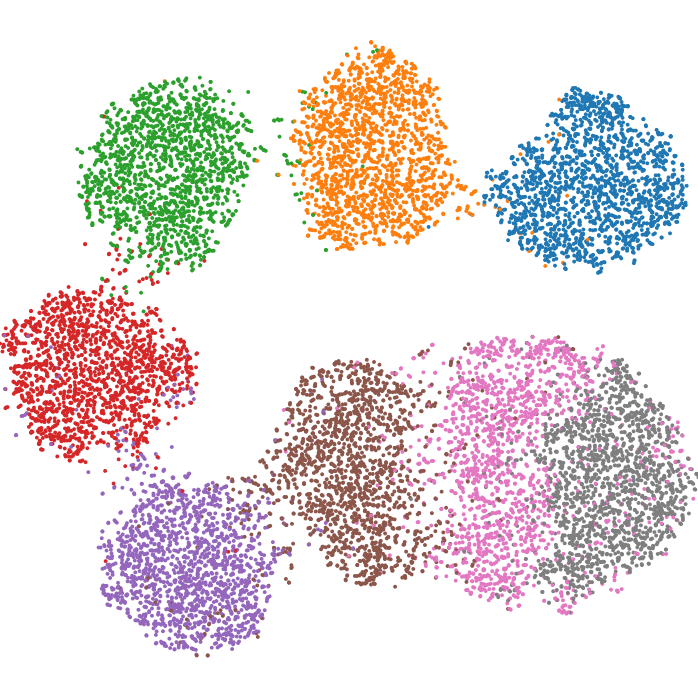} & 
    \includegraphics[width=\tSNEPanelWidth]{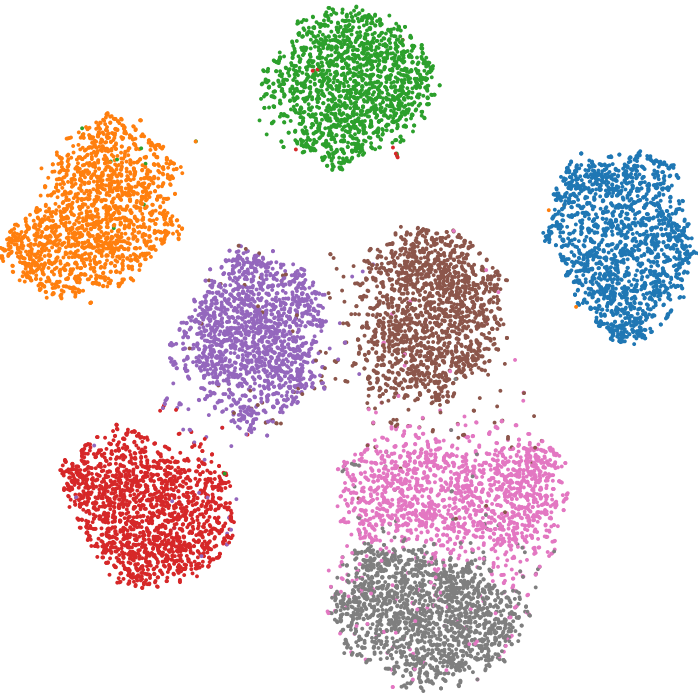} & 
    \includegraphics[width=\tSNEPanelWidth]{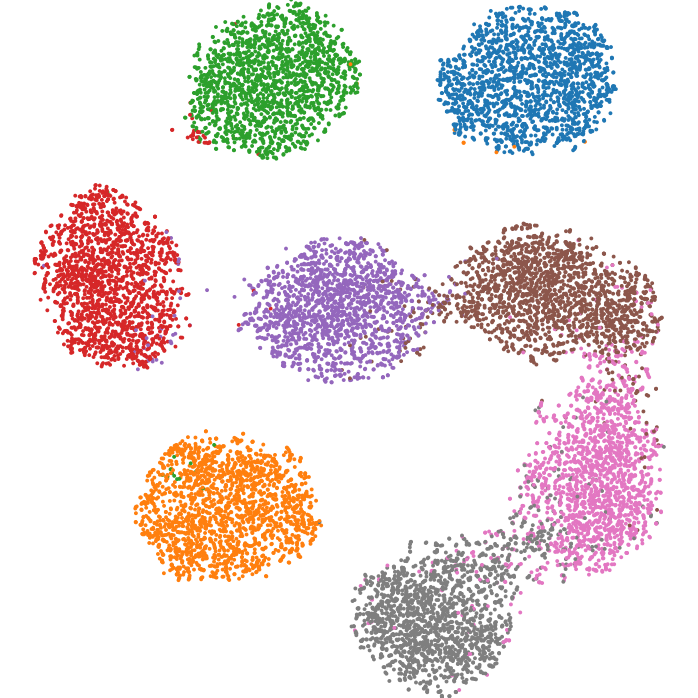} & 
    \includegraphics[width=\tSNEPanelWidth]{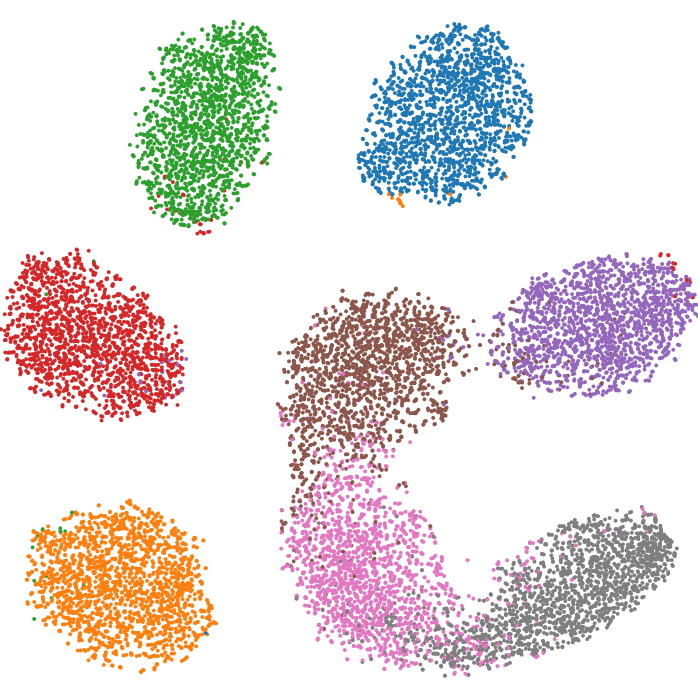} \\
\end{tabular}

%% file: tables/tsne/eurosat.tex
\setlength{\tabcolsep}{\tSNEColSpace}
\begin{tabular}{@{} ccccc @{}}
    \multicolumn{4}{c}{\footnotesize{VTAB EuroSAT}} \\
    \addlinespace[0.1cm]
    \tSNETitle{\finetune (97.1)} &
    \tSNETitle{\linear  (93.8)} & 
    \tSNETitle{\bias  (96.7)} & 
    \tSNETitle{\vptdeep  (95.7)} 
    \\
    \includegraphics[width=\tSNEPanelWidth]{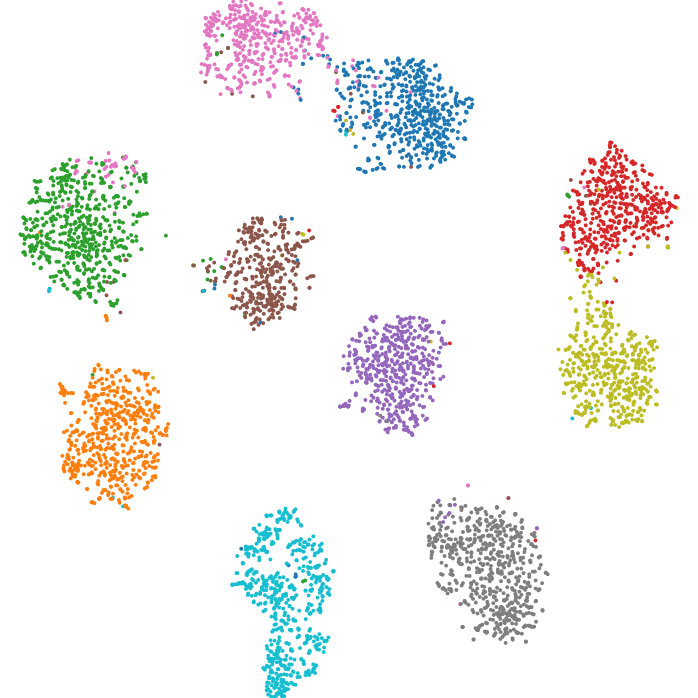} & 
    \includegraphics[width=\tSNEPanelWidth]{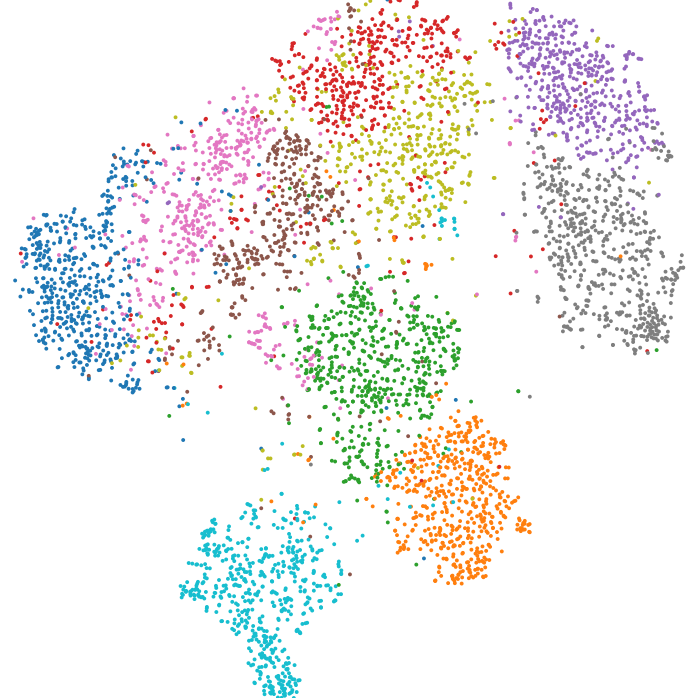} & 
    \includegraphics[width=\tSNEPanelWidth]{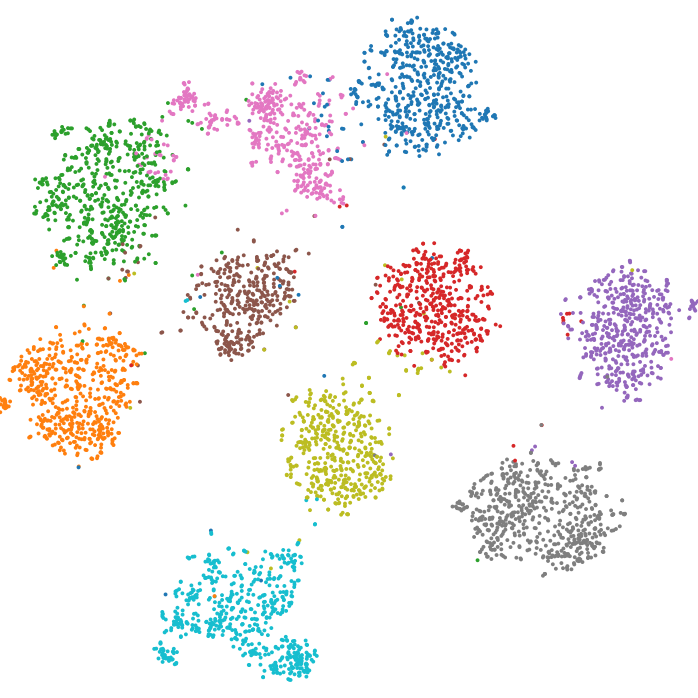} & 
    \includegraphics[width=\tSNEPanelWidth]{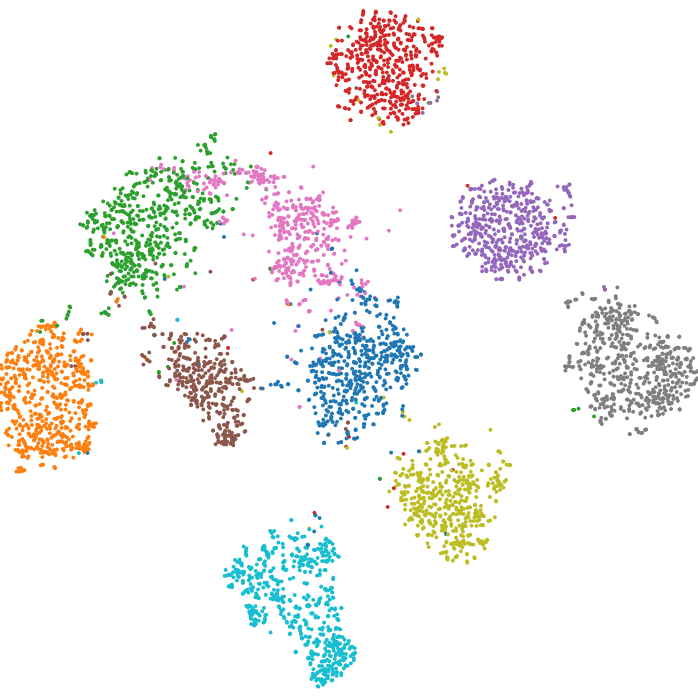} 
    \\
    \tSNETitle{\adaptformer  (96.9)} & 
    \tSNETitle{\ssf  (96.5)} & 
    \tSNETitle{\lora (96.7)} & 
    \tSNETitle{\apla  (97.6)} 
    \\
    \includegraphics[width=\tSNEPanelWidth]{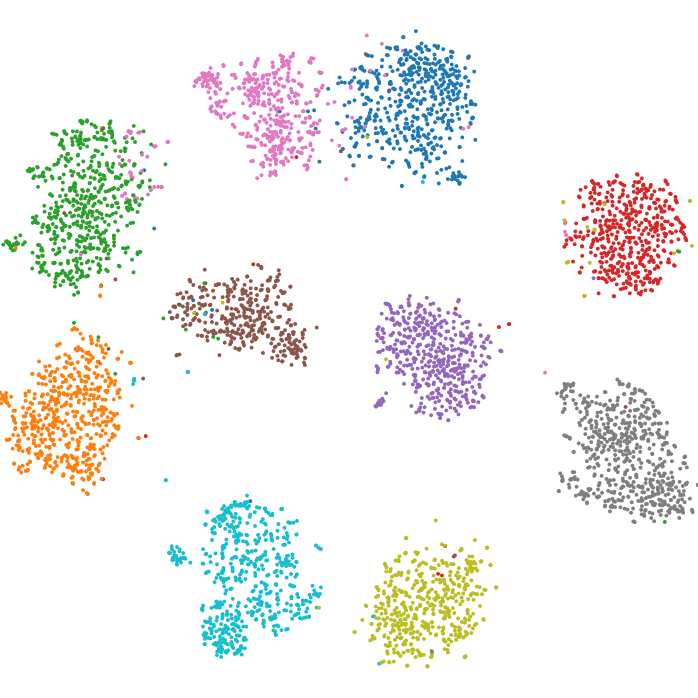} & 
    \includegraphics[width=\tSNEPanelWidth]{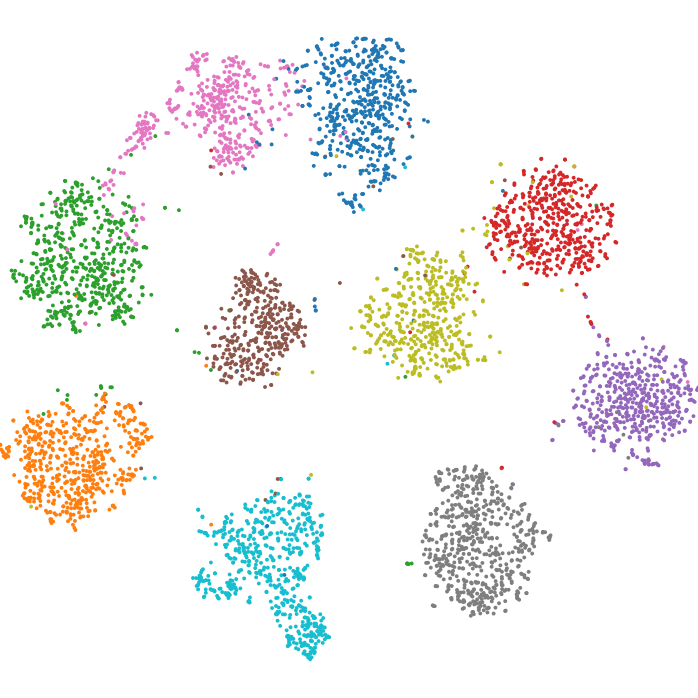} & 
    \includegraphics[width=\tSNEPanelWidth]{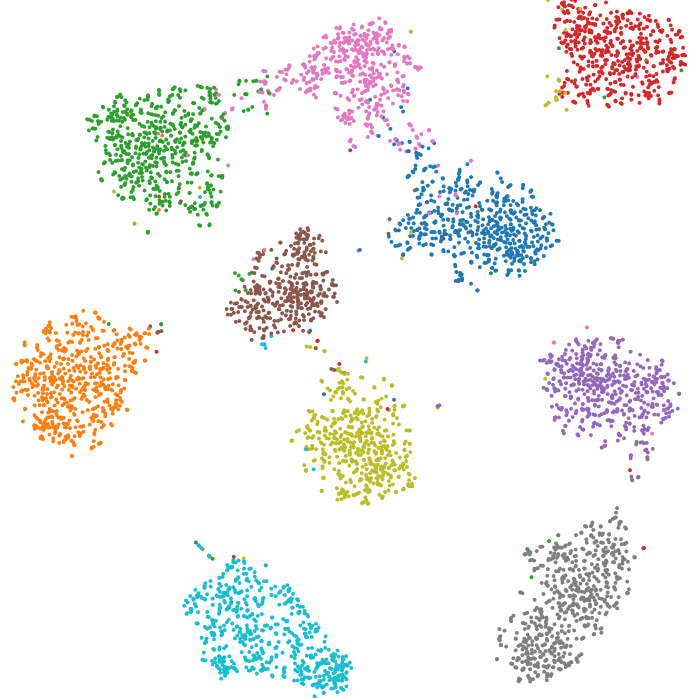} & 
    \includegraphics[width=\tSNEPanelWidth]{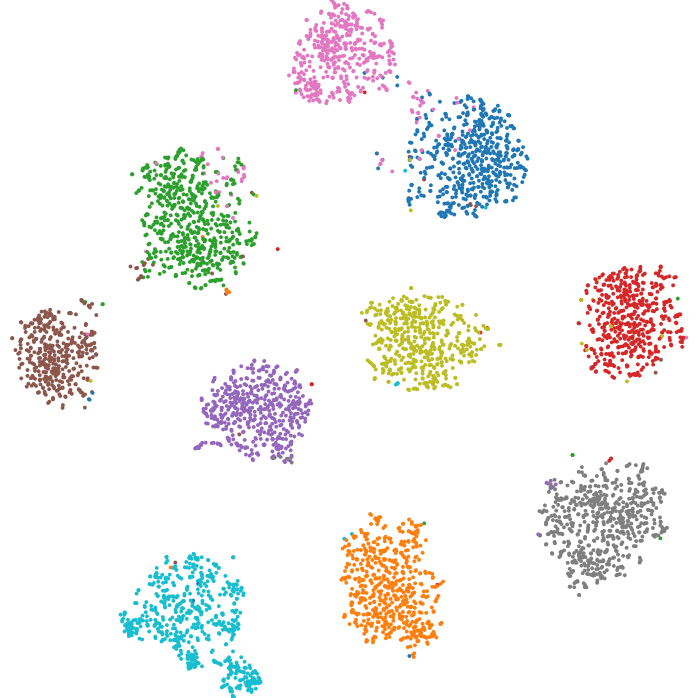} \\
\end{tabular}